\documentclass[10pt,journal,compsoc]{IEEEtran}
\UseRawInputEncoding

\usepackage{cite}
\usepackage{multirow}
\usepackage{amsmath,amssymb,amsfonts}
\usepackage{algorithmic}
\usepackage{verbatim}
\usepackage{diagbox}
\usepackage[ruled]{algorithm2e}
\usepackage{graphicx}
\usepackage{textcomp}
\usepackage{xcolor}
\usepackage{subfigure}
\usepackage{epsfig,epstopdf}

\begin{document}

\title{Crowdsourcing with Meta-Workers: A New Way to Save the Budget}

\author{Guangyang Han,
        Guoxian Yu,
        Lizhen Cui,
        Carlotta Domeniconi,
        Xiangliang Zhang

\IEEEcompsocitemizethanks{
\IEEEcompsocthanksitem G. Han is with the College of Computer and Information Sciences, Southwest University, Chongqing 400715, China. \protect\\
E-mail: hanguangyang@email.swu.edu.cn.

\IEEEcompsocthanksitem G. Yu and L. Cui are with the School of Software at Shandong University, Jinan 250101, China. G. Yu is the corresponding author. \protect\\
E-mail: \{gxyu, clz\}@sdu.edu.cn.

\IEEEcompsocthanksitem C. Domeniconi is with the Department of Computer Science, George Mason University, VA 22030, USA. \protect\\
E-mail: carlotta@cs.gmu.edu.

\IEEEcompsocthanksitem X. Zhang is with the Department of Computer Science, King Abdullah University of Science and Technology, Saudi Arabia. \protect\\
E-mail: xiangliang.zhang@kaust.edu.sa.
}}

\markboth{IEEE Class File}%
{Shell \MakeLowercase{\textit{et al.}}: Bare Advanced Demo of IEEEtran.cls for IEEE Computer Society Journals}

\IEEEtitleabstractindextext{
\begin{abstract}
Due to the unreliability of Internet workers, it's difficult to complete a crowdsourcing project satisfactorily, especially when the tasks are multiple and the budget is limited. Recently, meta learning has brought new vitality to few-shot learning, making it possible to obtain a classifier with a fair performance using only a few training samples. Here we introduce the concept of \emph{meta-worker}, a machine annotator trained by meta learning for types of tasks (i.e., image classification) that are well-fit for AI. Unlike regular crowd workers, meta-workers can be reliable, stable, and more importantly, tireless and free. We first cluster unlabeled data and ask crowd workers to repeatedly annotate the instances nearby the cluster centers; we then leverage the annotated data and meta-training datasets to build a cluster of meta-workers using different meta learning algorithms. Subsequently, meta-workers are asked to annotate the remaining crowdsourced tasks. The Jensen-Shannon divergence is used to measure the disagreement among the annotations provided by the meta-workers, which determines whether or not crowd workers should be invited for further annotation of the same task. Finally, we model meta-workers' preferences and compute the consensus annotation by weighted majority voting. Our empirical study confirms that, by combining machine and human intelligence, we can accomplish a crowdsourcing project with a lower budget than state-of-the-art task assignment methods, while achieving a superior or comparable quality.
\end{abstract}

}

\maketitle

\IEEEdisplaynontitleabstractindextext

\IEEEpeerreviewmaketitle

\section{Introduction}
\IEEEPARstart{C}{rowdsourcing} used to refer to the practice of an organization of outsourcing tasks, otherwise performed by employees, to a large number of volunteers \cite{howe2006rise}. Recently, crowdsourcing has become the only viable way to annotate massive data through the hiring of a large number of inexpensive Internet workers \cite{sheng2019machine}. Although a variety of tasks can be crowdsourced, the relatively common one is the annotation of images (i.e., ImageNet) for data-driven machine learning algorithms such as deep learning. However, due to the difficulty of the tasks, the poor task description, and the diverse capacity ranges of workers and so on \cite{li2016crowdsourced, chittilappilly2016survey}, we often need to invite multiple workers to annotate the same data to improve the label quality \cite{sheng2008get}. This limits the use of crowdsourcing when the available budget is limited. As such, we face the need of obtaining quality data with a tight budget. Proposed solutions focus on modeling crowdsourced tasks \cite{zheng2016docs,boim2012asking}, workers \cite{welinder2010multidimensional}, or crowdsourcing processes \cite{zheng2015qasca,li2016crowdsourcing,tu2020attention} to achieve a better understanding of the same, thereby reducing the impact of incompetent worker and the number of repeated annotations, while improving the quality.

We argue that meta learning can offer a solution to the challenge we are facing with crowdsourcing \cite{vanschoren2018meta}. Meta learning imitates the process of learning in humans. With only a few data in the target domain, the learner can quickly adapt to recognize a new class of objects. For example, state-of-the-art meta learning algorithms can achieve an accuracy of nearly 60\% on five classification tasks for the Mini-ImageNet dataset, with only one training instance per class (called 5-way 1-shot few-shot learning) \cite{sung2018learning}. This is comparable to the capability of the majority of human workers on real world crowdsourcing platforms\cite{kazai2011worker}. As such, we can model crowdsourcing employees as meta learners: with a small amount of guidance, they can quickly learn new skills to accomplish crowdsourced tasks.

To make this idea concrete, we introduce the notion of \textit{meta-worker}, a virtual worker trained via a meta learning algorithm, which can quickly generalize to new tasks. Specifically, our crowdsourcing process is formulated as follows. Given a crowdsourcing project, we first partition the tasks into different clusters. We then collect a batch of tasks close to each cluster center and ask the crowd workers to annotate them until a `$n$-way $k$-shot' meta-test dataset is obtained. We also build our meta-training datasets by collecting data from the Internet. Different meta learning algorithms are used to generate a group of diverse meta-workers. We then employ the meta-workers to annotate the remaining tasks; we measure the disagreement among the annotations using the Jensen-Shannon divergence, and consequently decide whether or not to further invite crowd workers to provide additional annotations. Finally, we model the meta-workers' preference and use weighted majority voting (WMV) to compute the consensus labels, and iteratively optimize the latter until convergence is reached.

The main contributions of our work are as follows:\\
(i) This work is the first effort in the literature to directly supplement human workers with machine classifiers for crowdsourcing. The results indicate that machine intelligence can limit the use of crowd workers and achieve quality control.\\
(ii) We use meta-learning to train meta-workers. In addition, we employ ensemble learning to boost the meta-workers' ability of producing reliable labels. Most simple tasks do not require the participation of human workers, thus enabling budget savings.\\
(iii) Experiments on real datasets prove that our method achieves the  highest quality while using comparable or far less budget than  state-of-the-art methods, and the amount of budget-saving grows as the scale of tasks increases.

\section{Related Work}
Our work is mainly related to two research areas, crowdsourcing and meta learning. Meta learning (or learning to learn) is inspired by the ability of humans to use previous experience to quickly learn new skills \cite{vanschoren2018meta,liu2020FSLHierarchy}. The meta learning paradigm trains a model using a large amount of data in different source domains where data are available, and then fine tunes the model using a small number of samples of the target domain.

In recent years, a variety of meta learning approaches have been introduced. Few-shot meta learning methods can be roughly grouped into three categories, namely optimization-based, model-based (or memory-based, black box model), and metric-based (or non-parametric) methods. Optimization-based methods treat meta learning as an optimization problem and extract the meta-knowledge required to improve the optimization performance. For example, Model-Agnostic Meta-Learning \cite{finn2017model} looks for a set of initialization values of the model parameters that can lead to a strong generalization capability. The idea is to enable the model to quickly adapt to new tasks using few training instances. Model-based methods train a neural network to predict the output based on the model's current state (determined by the training set) and input data. The meta-knowledge extraction and meta learning process are wrapped in the model training process. Memory-Augmented Neural Networks \cite{santoro2016meta} and Meta Networks \cite{munkhdalai2017meta} are representative model-based methods. Metric-based methods use clustering ideas for classification. They perform non-parametric learning at the inner (task) level by simply comparing validation points with training points, and assign a validation point to the category with the closest training point. This approach has several representative methods: Siamese networks \cite{bromley1993signature}, prototypical networks \cite{snell2017prototypical} and relation networks \cite{sung2018learning}. Here the meta-knowledge is given by the distance or similarity metric.

Few-shot meta learning can leverage multi-source data to better mine the information of the target domain data, which is consistent with the aim of budget saving in crowdsourcing. A lot of effort has been put to achieve budget savings in crowdsourcing \cite{li2017crowdsourced}. One way is to reduce the number of tasks to be performed, such as task pruning (prune the tasks that machines can do well) \cite{wang2012crowder}, answer deduction (prune the tasks whose answers can be deduced from existing crowdsourced tasks) \cite{wang2013leveraging}, and task selection (select the most beneficial tasks to crowdsource) \cite{mozafari2014scaling,yoo2019learning,tu2020crowdwt,yu2020active}. Another approach is to reduce the cost of individual tasks, or dynamically determine the price of single task, mainly by better task (flow) design \cite{marcus2012counting,zheng2018dlta,tong2018dynamic,tong2018slade}.

Our work is also related to semi-supervised learning. Semi-supervised self-training \cite{yarowsky1995unsupervised,xie2020self} gradually augments the labeled data with new instances, whose labels have been inferred with high confidence, until the unlabeled data pool is empty or the learner does not improve any further. The effectiveness of this approach depends on the added value of the augmented labeled data. Furthermore, the model needs to be updated every time when an instance is augmented as labeled, which is not feasible in a large crowdsourcing project. Some active learning based crowdsourcing approaches \cite{fang2014active,korycki2017combining,yu2020active} also suffer from these issues. \cite{zhang2017improving} trained a group of classifiers using cleaned data in crowdsourcing, the classifiers are then used to correct the potential noisy labels. Unlike the above proposed solutions, our approach is feasible because we directly use meta-workers, trained by meta learning, to annotate unlabeled data. Meta-workers can quickly generalize to new tasks and can achieve a good performance with the support of only few instances. In contrast, existing methods canonically depend on sufficient training instances for each category to enable machine intelligence assisted crowdsourcing \cite{korycki2017combining,xie2020self,zhang2017improving,zhang2018ensemble}. In addition, our model learns meta-knowledge from external free data to save the budget considerably. Due to the cooperation among diverse meta-workers and to ensemble learning, we can further boost the performance of a group of meta-workers, without the need of frequent updates.

\section{Proposed Methodology}

\subsection{Definitions}
\begin{figure}[h!tbp]
\centerline{\includegraphics[width=9cm]{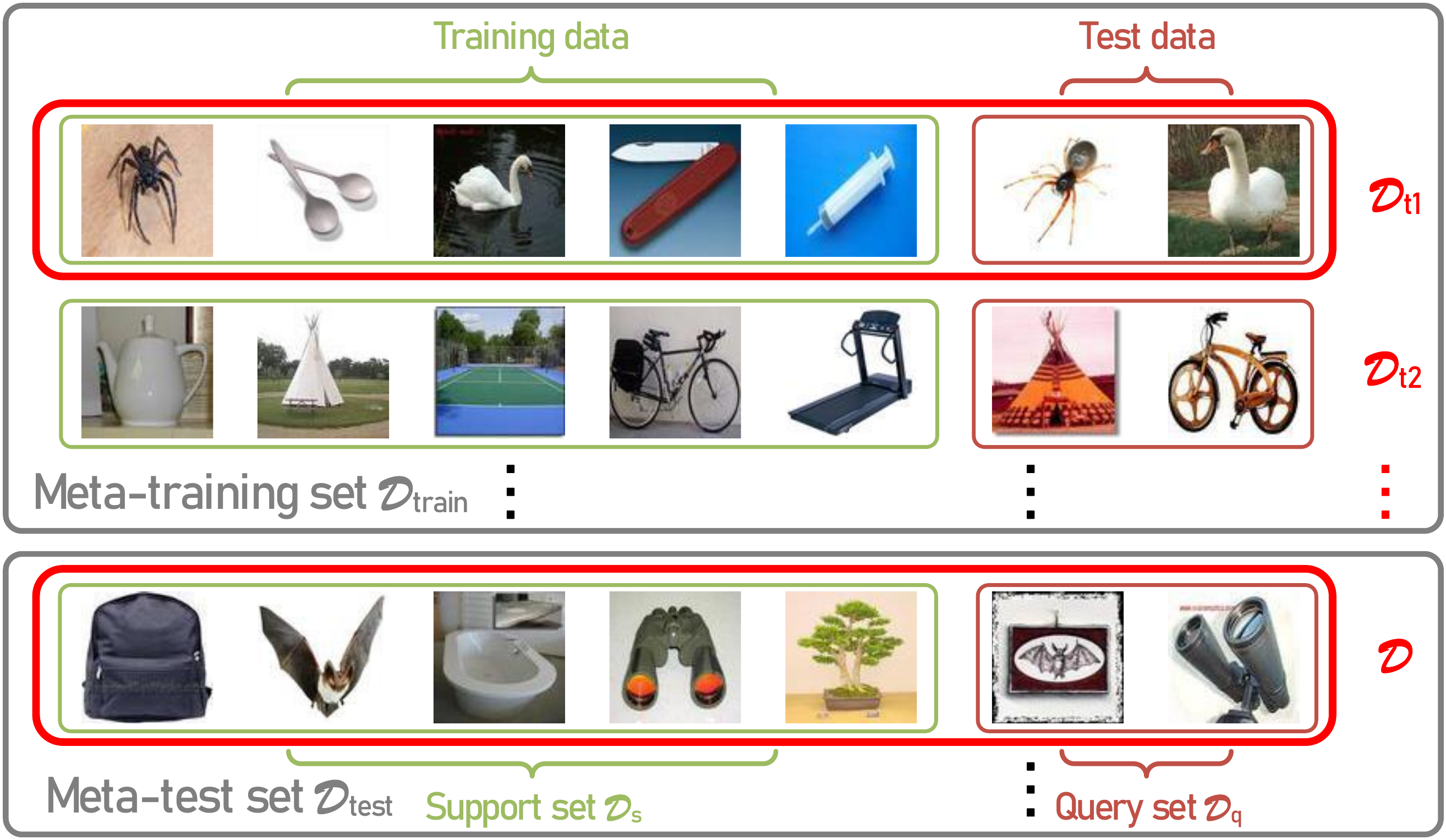}}
\caption{An example of 5-way 1-shot meta learning dataset. From the perspective of meta learner, data can be divided into meta-training set $\mathcal{D}_{train}$ and meta-test set $\mathcal{D}_{test}$. The former is composed of datasets of many independent tasks $\mathcal{D}_{ti}$, and the latter can be separated as support set $\mathcal{D}_s$ and query set $\mathcal{D}_q$. The label space size is called \textit{n-way}, and the number of instance per label in $\mathcal{D}_s$ is called \textit{k-shot}.}
\label{metadata}

\end{figure}

In this section, we will formalize the Crowdsourcing with Meta-Workers problem setup in detail. Meta model usually consists of two learners, the upper one is called `meta learner', with the duty of extracting meta-knowledge to guide the optimization of the bottom learner, and the bottom one is called `base learner', which executes the classification job. In order to achieve this, the model is firstly trained on a group of different machine learning tasks (like Multi-task Learning, MTL \cite{zhang2018mtl}) named ``meta-training set", expecting the model to be capable for different tasks, then the model moves to its target domain called ``meta-test set". More precisely, the meta learner takes one classical machine learning task as a meta-training instance, and a group of tasks as meta-training set, extracts meta-knowledge from them, then uses these meta-knowledge to guide the training and generalization of the base learner on the target domain. To eliminate ambiguity, following the general naming rules of meta learning, we use `train/test' to distinguish the instance (classic machine learning task) used by meta learner, and `support/query' to distinguish the instance (instance in classical machine learning) used by base learner, the composition of dataset required for meta learning is shown in Fig. \ref{metadata}.

Let $\mathcal{D} = \{x_1, x_2, \cdots, x_N\}$ be a crowdsourcing project with $N$ tasks, where each task $x_i$ belongs to one out of $n$ classes. We cluster the $N$ tasks into $n$ categories, and select $k$ instances from each cluster to be annotated by crowd workers. The resulting $n \times k$ annotated tasks $\mathcal{D}_s$ ($s$-support set, $q$-query set) are used to estimate the crowd workers' capacity. $\mathcal{D}_s$ and the remaining tasks $\mathcal{D}_q = \mathcal{D} - \mathcal{D}_s$ form our meta-test dataset: $\mathcal{D}_{test} = \mathcal{D}_s \cup \mathcal{D}_q = \sum_{i=1}^{nk}\{x_i, y_i\} \cup \sum_{i=1}^{N-nk}\{x_i\}$. We also need to collect auxiliary datasets related to the $N$ tasks at hand to build our meta-training dataset $\mathcal{D}_{train} = \{\mathcal{D}_{t1}, \mathcal{D}_{t2}, \cdots\}$, where each $\mathcal{D}_{ti}$ is an independent machine learning task dataset. The diversity of $\mathcal{D}_{train}$ guarantees the generalization ability. In the few-shot learning paradigm, this setup is called a `$n$-way $k$-shot' problem. Table \ref{marks} summarizes the notations used in this paper.

\begin{table}[h!tbp]
\caption{List of symbols.}
\begin{center}
\begin{tabular}{c|c|c}
\hline
\textbf{Item}           & \textbf{Symbol}               & \textbf{Remarks}            \\ \hline
number of crowd task    & $N$                           & total $N$ tasks            \\ \hline
task / instance         & $x$                           & total $N$ tasks            \\ \hline
task labels' vector     & $\mathbf y$                   & size $N$, value [1, 2, $\cdots$, $n$]   \\ \hline
label space size        & $n$                           & called $n$-way            \\ \hline
support instance num    & $k$                           & called $k$-shot            \\ \hline
meta-test task type     & $s\ /\ q$                     & support / query             \\ \hline
worker (set)            & $w\ (\mathcal{W})$            & total $W_m$ meta, $W_c$ crowd   \\ \hline
worker type             & $m\ /\ c$                     & meta / crowd     \\ \hline
confusion matrix        & $\mathbf M$                   & size $n \times n$, model meta    \\ \hline
accuracy / capacity     & $\mu$                         & decimal, model crowd        \\ \hline
worker's annotations    & $\mathbf{A}$                  & size $N \times n$ \\ \hline
annotation              & $\mathbf a$                   & size $n$   \\ \hline
meta algorithms (set)   & $f\ (\mathcal F)$             & total $W_m$ algorithms        \\ \hline
divergence threshold    & $\theta$                      & difficulty criteria              \\ \hline
\end{tabular}
\label{marks}
\end{center}
\end{table}

\subsection{Workflow}
\begin{figure}[h!tbp]
\centerline{\includegraphics[width=9cm]{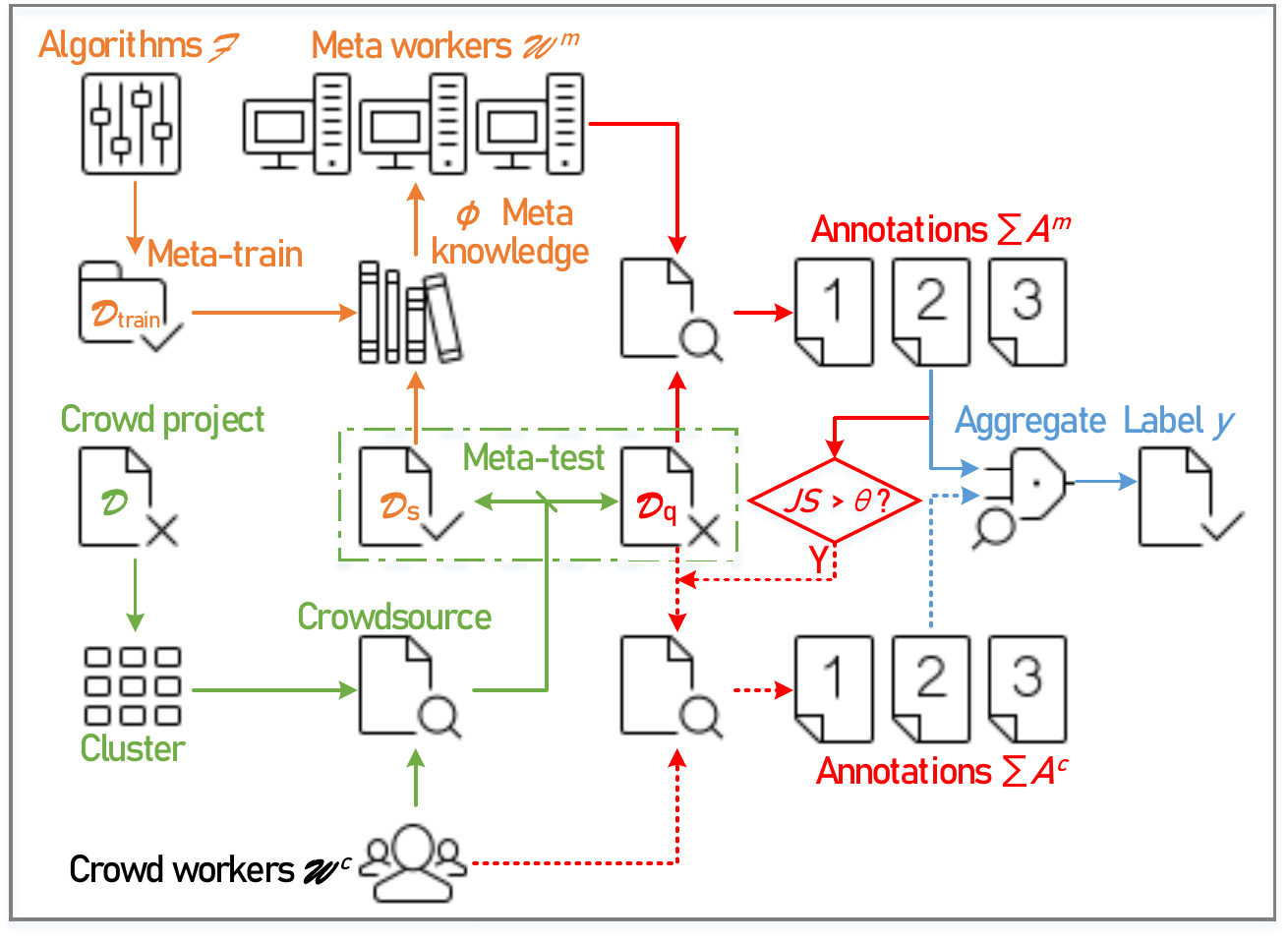}}
\caption{Workflow of our approach. Four steps are represented using different colors. The meta-test set (dash box) is built from the project $\mathcal{D}$ via clustering and crowdsourcing. $\mathcal{D}_s$, together with the meta-knowledge $\phi$, generates meta-workers $\mathcal{W}^m$; the latter annotates $\mathcal{D}_q$ to attain the matrices $\{\mathbf{A}^m\}_{m=1}^{W_m}$. If the meta-workers disagree on task $x_i$ (i.e. $JS > \theta$), crowd workers will be asked to provide annotations (red dash line). Finally, we aggregate $\{\mathbf{A}^m\}_{m=1}^{W_m} \text{ and } \{\mathbf{A}^c\}_{c=1}^{W_c}$ to acquire the labels.}
\label{flowchart}
\end{figure}

The workflow of our approach is shown in Fig. \ref{flowchart}. After obtaining $\mathcal{D}_{train}$ (completely labeled) and $\mathcal{D}_{test}$ (partially labeled), the problem becomes a standard `$n$-way $k$-shot' meta learning problem. We apply a meta learning algorithm $f \in \mathcal{F}$ on $\mathcal{D}_{train}$ to extract the meta-knowledge $\phi = f(\mathcal{D}_{train})$. We then combine $\phi$ with the meta-test support set $\mathcal{D}_s$ to adjust $\phi$ to the target task domain, obtaining a meta-worker (i.e., a classifier) $w^m$. By using $W_m$ different meta learning algorithms in $\mathcal{F}$, we can obtain a group of meta-workers $\mathcal{W}^m = \{w_1^m, w_2^m, \cdots, w_{W_m}^m\}$ with different preferences. $\mathcal{W}^m$ are used to get the annotation matrices $ \{\mathbf{A}^m\}_{m=1}^{W_m}$ of the remaining tasks $\mathcal{D}_q$. If the meta-workers disagree with one another on a task, we invite crowd workers to provide further annotations for that task. Finally, we use the confusion matrix and the accuracy to separately model the preferences of meta-workers and of crowd workers, and compute the consensus labels by weighted majority voting in an iterative manner.

\subsection{Building the Meta-test Set}
The first step of our method transforms a crowdsourcing project of $N$ tasks into an $n$-way $k$-shot meta-test set $\mathcal{D}_{test} = \mathcal{D}_s \cup \mathcal{D}_q$. In order to build an $n$-way $k$-shot meta-test support set $\mathcal{D}_s$ from $N$ unlabeled instances, we use $K$-means to cluster the instances into $n$ clusters (other clustering algorithms can be used as well). We then select $k$ instances closest to each cluster center to be annotated. Since the results of clustering are not perfect, the selected instances might not belong to the assigned cluster. Therefore, building $\mathcal{D}_s$ requires slightly more than $n \times k$ instances.

The label quality of the meta-test support set $\mathcal{D}_s$ is crucial. As such, we ask as many as possible workers ($\leq 30$) to provide annotations. A basic assumption in crowdsourcing is that the aggregated annotations given by a large number of workers are reliable. For example, with an average accuracy of crowd workers as 0.6 under 5 classes, even if the simplest majority voting is used, the expected accuracy of 10 repeated annotations from 10 crowd workers is about 95\%, and of 30 repeated annotations is above 99.95\%. Once $\mathcal{D}_s$ is attained, the remaining tasks of $\mathcal{D}$ give $\mathcal{D}_q$. $\mathcal{D}_s$, combined with $\mathcal{D}_q$, constitutes the meta-test set.

{Although the ground truth of each task in crowdsourcing is unknown and it is hard to estimate a worker’s quality, we can still approximate the ground truth for a small portion of tasks (called golden tasks) and use them to estimate worker's quality \cite{zheng2017truth,tu2020crowdwt}.
In this way, we can pre-identify low-quality workers based on $\mathcal{D}_s$, and prevent them from participating into the subsequent crowdsourcing process. The modeling of crowd worker's quality will be discussed in detail in Section \ref{Aggregeta}.}

\subsection{Training Meta-workers}
We need to build meta-workers using meta learning algorithms. We choose one representative method from each of the three meta learning categories to form our meta-worker cluster, namely Model-Augmented Meta-Learning (MAML) \cite{finn2017model}, Meta Networks (MN) \cite{munkhdalai2017meta}, and Relation Networks (RN) \cite{sung2018learning}, with a $5$-way $5$-shot setting. Each induces a different learning bias, thus together they can lead to effective ensembles.

Meta learning trains a model on a variety of learning tasks. The model is then fine-tuned to solve new learning tasks using only a small number of training samples of the target task domain \cite{vanschoren2018meta}. The general meta-learning algorithm can be formalized as follows:

\begin{align}
    \log p(\pi | \mathcal{D}_s,\mathcal{D}_{train}) &= \log\int_\Phi p(\pi | \mathcal{D}_s,\phi)p(\phi | \mathcal{D}_{train}) d\phi \notag
    \\ &\approx \log p(\pi | \mathcal{D}_s, \phi^*) + \log p(\phi^* | \mathcal{D}_{train}) \notag
    \\ \textrm{Here}:\ \ \phi^* &= \arg \max_\phi \log p(\phi | \mathcal{D}_{train}) \label{learning}
    \\ \arg\max_\pi \log p(\pi | \mathcal{D}_s &,\mathcal{D}_{train}) \approx \arg\max_\pi \log p(\pi | \mathcal{D}_s, \phi^*) \label{meta}
\end{align}
where $\pi$ represents the model parameters we want to learn, and $\phi^*$ is the meta-knowledge extracted from $\mathcal{D}_{train}$. Eq. (\ref{learning}) corresponds to the meta-knowledge learning phase and Eq. (\ref{meta}) comes to the meta-knowledge adaptation phase.

For the above purpose,
\emph{MAML} optimizes the initial values of the parameters to enable the model to adapt quickly to new tasks during gradient descent. The meta-knowledge $\phi_{\text{MAML}}$ is given by the gradient descent direction.
\emph{MN} has the ability to remember old data and to assimilate new data quickly. \emph{MN} learns meta-level knowledge across tasks and shifts its inductive biases along the direction of error reduction.
\emph{RN} learns a deep distance metric ($\phi_{\text{RN}}$) to compare a small number of instances within episodes.

\subsection{Obtaining Annotations}
After the meta-workers $\mathcal{W}^m$ have been adjusted to the target task domain, they can be used to replace, or work together with, the crowd workers to provide the annotations $\mathbf{A}^m$ and $\mathbf{A}^c$ for $\mathcal{D}_q$, and to save the budget. Although we consider multiple meta-workers to improve the quality of crowdsourcing, there may exist difficult tasks that cannot be performed well by meta-workers. Therefore, we model the difficulty of all the tasks, and select the difficult ones to be annotated by crowd workers.

There are many criteria to quantify the difficulty of a task \cite{li2017crowdsourced,tu2020crowdwt,yu2020active}. Here, we adopt a simple and intuitive criterion: the more difficult a task is, the harder is for meta-workers to reach an agreement on it, and the larger the divergence between the task annotations is. As such, we can approximate the difficulty of a task by measuring the annotation divergence. Since the annotation given by meta-workers is a label probability distribution, the KL divergence (Eq. (\ref{KL})) can be used to measure the difference between any two distributions:

\begin{equation}
    KL(\mathbf{P}||\mathbf{Q}) = \sum \mathbf{P}(x)\log\frac{\mathbf{P}(x)}{\mathbf{Q}(x)}
    \label{KL}
\end{equation}
where $\mathbf{P}$ and $\mathbf{Q}$ are discrete probability distributions.

However, the direct use of the KL divergence has two disadvantages, i.e., the KL divergence is asymmetric and it can only calculate the divergence between two annotations. Asymmetry makes it necessary to consider the order between annotations, which makes it more complicated and tedious to measure the divergence of multiple annotations. For those reasons, we use the symmetric Jensen-Shannon divergence (Eq. (\ref{JS})), an extension of the KL divergence, to measure the divergence of all possible annotation pairs, and take their average value as the final divergence (Eq. (\ref{mJS})):

\begin{align}
    JS(\mathbf{P}||\mathbf{Q}) = \frac{1}{2}(KL(\mathbf{P}(x)||\mathbf{R}(x)) + KL(\mathbf{Q}(x)||\mathbf{R}(x)))  \label{JS} \\
    \textrm{Here}:\ \ \mathbf{R}(x) = \frac{\mathbf{P}(x) + \mathbf{Q}(x)}{2}  \ \ \ \ \ \ \ \ \ \ \ \ \ \ \ \ \ \ \ \ \notag
\end{align}

\begin{equation}
    JS(\mathcal{A})= C_{|\mathcal{A}|}^2 \sum\limits_{1 \leq i < j \leq |\mathcal{A}|}JS(\mathcal{A}_i||\mathcal{A}_j)
    \label{mJS}
\end{equation}
where $\mathcal{A}$ represents the set of meta annotations for a task.

Once we have collected the meta annotations $\{\mathbf{A}^m\}_{m=1}^{W_m}$, we calculate the JS divergence of each task, pick out the tasks with divergence greater than $\theta$, and submit them to crowd workers for further annotation. We assign additional ${W_m}$ crowd workers with fair quality to the difficult tasks each, and obtain the crowd annotations $\{\mathbf{A}^c\}_{c=1}^{W_c}$. 
Finally, all annotations $\{\mathbf{A}^m\}_{m=1}^{W_m} \cup \{\mathbf{A}^c\}_{c=1}^{W_c}$ are gathered to compute the consensus labels.

\subsection{Aggregating the Annotations} \label{Aggregeta}

\subsubsection{Correcting Annotations}
During the last step we compute the consensus labels of the tasks. Meta annotations and crowd annotations are inherently different: the former is a discrete probability distribution in the label space, while the crowd annotation is a typical one-hot coding in the label space. Therefore, we use different strategies to model meta-workers and crowd workers. The meta-workers' probability distribution annotation gives the probabilities that the instance belongs to each class, which are suitable for a D\&S model \cite{dawid1979maximum,whitehill2009whose,ipeirotis2010quality}. The one-hot coding crowd annotations, instead, simply indicate the chosen most likely label for an instance. Furthermore, the number of crowd annotations is smaller than that of meta annotations. As such, we can't build a complex model for crowd workers based on their annotations, so we choose the simplest but effective Worker Probability model (or capacity, accuracy etc.) \cite{guo2012so,karger2011iterative,zhang2013reducing} for modeling in this case.

Since not all the tasks are annotated by crowd workers and $\mathbf{A}^c$ is incomplete, we use negative `dummy' annotations $\mathbf{a}^c_{pad} = [-1,-1, \cdots, -1]$ to fill up $\mathbf{A}^c$. To eliminate the difference between meta annotations and crowd ones, and to model crowd workers, we introduce an accuracy value $\mu$ for each crowd worker, and transform a crowd worker's annotation as follows:
\begin{equation}
    \widetilde{\mathbf a}^c_i = Tr(\mathbf{a}^c | \mu) =
    \begin{cases}
    \frac{1 - \mu}{n - 1} & \mathbf{a}^c_i = 0 \\
    \mu & \mathbf{a}^c_i = 1\\
    0 & \textrm{otherwise}
    \end{cases} \ \textrm{for\ $i\in \{1, \cdots, n\}$}
    \label{mu}
\end{equation}
We perform the above transformation on each bit of the annotation vector $\mathbf a^c$, each $\mu$ of the crowd worker is initialized when $\mathcal{D}_s$ is attained.

The D\&S model focuses on single-label tasks (with fixed $n$ choices) and models the bias of each meta-worker as a confusion matrix $\mathbf{M}^w$ with size $n \times n$. $\mathbf{M}_{p,q}^w$ in $\mathbf{M}^w$ models the probability that worker $w$ wrongly assigns label $\mathbf{y}_q$ to an instance of true label $\mathbf{y}_p$. We use the confusion matrix $\mathbf{M}^w$ of a meta-worker $w$ to correct the annotation results $\mathbf{A}^w$ using Eq. (\ref{DS}), where $\widetilde{\mathbf{A}}^w$ is the corrected annotation. We initialize each $\mathbf M^w$ with an identity matrix of size $n \times n$ in the first iteration.


\begin{align}
    \mathbf{A}^w = \widetilde{\mathbf A}^w \cdot \mathbf{M}^w \ \  \Rightarrow\ \ \widetilde{\mathbf A}^w = \mathbf{A}^w \cdot (\mathbf{M}^w)^{-1}
    \label{DS}
\end{align}

\subsubsection{Inferring Labels}
Once the above correction process is performed via Eqs. (\ref{mu}) and (\ref{DS}), we compute the consensus labels using weighted majority voting on the corrected annotations. We then use the inferred labels to update the confusion matrix $\{\mathbf{M}^w\}_{w=1}^{W_m}$ of the meta-workers and the accuracy values $\{\mu^w\}_{w=1}^{W_c}$ of the crowd workers. Let $\Psi = \{\{\mathbf{M}^w\}_{w=1}^{W_m},\ \{\mu^w\}_{w=1}^{W_c}\}$. We use the \textit{EM} algorithm \cite{dawid1979maximum,ipeirotis2010quality,wang2012crowder} to optimize $\Psi$ and consensus labels $\mathbf y$ until convergence. The detailed process is as follows.

\textbf{E-step:} We use Eqs. (\ref{mu}) and (\ref{DS}), and $\Psi$ to correct $\{\mathbf{A}^m\}_{m=1}^{W_m}$ and $\{\mathbf{A}^c\}_{c=1}^{W_c}$. Then we combine $\{\widetilde{\mathbf A}^m\}_{m=1}^{W_m}$ and $\{\widetilde{\mathbf A}^c\}_{c=1}^{W_c}$ at the task level to obtain a larger annotation tensor $\mathbf T$ of size $N \times ({W_m}+{W_c}) \times n$, where the first dimension is the number of tasks, the second is the number of workers, and each $\mathbf{T}_{i,j}$ is an annotation of size $n$. We sum all the $({W_m}+{W_c})$ annotations, task by task, and select the label corresponding to the position with the highest probability value as the ground truth, as described in Eq. (\ref{e-step}) (the annotations and $\hat{\mathbf a}^i$ are vectors of size $n$).

\begin{align}
    \{\hat{\mathbf a}_i\}_k &= {\sum}_{j=1}^{W_m+W_c} \mathbf{T}_{i,j,k}\ \ \ \textrm{for\ $k\in \{1, 2, \cdots, n\}$}  \notag \\
    \hat{\mathbf{y}}_i &= \arg\max_k \{\hat{\mathbf a}_i\}_k
    \label{e-step}
\end{align}


\textbf{M-step:} Here we use $\hat{\mathbf y}$ and $\{\mathbf{A}^m\}_{m=1}^{W_m}$, and $\{\mathbf{A}^c\}_{c=1}^{W_c}$ to update $\Psi$. A formal description is given in Eq. (\ref{m-step}). Eq. (\ref{count}) gives out {how to count} the number of tasks {correctly} answered by worker $w_j$, $\mathbb I(\cdot)$ is an indicator function such that $\mathbb I(A|B) = 1$ if $A$ is true under condition $B$, and $0$ otherwise. $\mathbf{A}^j$ is worker $w_j$'s corresponding annotation matrix.

\begin{equation}
    CNT(w_j) = {\sum}_{i=1}^{N} \mathbb I(\arg\max_p \{\mathbf{A}^j_i\}_p \ |\ \hat{\mathbf y}_i = p)
    \label{count}
\end{equation}

To update $\mu^j$ for crowd worker $w_j^c$, we first count the number of tasks that the worker has correctly annotated by Eq. (\ref{count}), and then normalize the count by the total number of annotations $N^j$ the worker has provided. Note that, in general, a crowd worker $w_j^c$ does not annotate all tasks (the negative dummy annotations $\mathbf{a}^c_{pad}$ are skipped).

We update the confusion matrix $\mathbf{M}^j$ of meta-worker $w_j^m$ row by row. Here $\mathbf{M}^j_{p,q}$ represents the probability of mistaking a task of label $\mathbf{y}_p$ as $\mathbf{y}_q$; as such, the denominator ($N_p$) is the number of tasks with label $\mathbf{y}_p$, and the numerator is the number of tasks whose label is $\mathbf{y}_p$, but is mistaken as $\mathbf{y}_q$. We need to count a total of $n \times n$ label confusion cases, and update each entry in the confusion matrix.

\begin{align}
    \hat{\mu}^j &= \frac{CNT(w_j^c)}{N^j} \notag \\
    \{\hat{\mathbf M}^j\}_{p,q} &= \frac{CNT(w_j^m)}{N_p}
    \label{m-step}
\end{align}

Our approach (\textbf{MetaCrowd}) is summarized in Algorithm \ref{alg1}. We first use clustering and crowdsourcing to transform crowdsourcing project into $n$-way $k$-shot few-shot learning problem (line 1-4). Then meta learning algorithms are invited to train meta-workers (line 5-10), which will annotate all the remaining crowdsourcing tasks. After that, JS divergence is employed to measure task's difficulties and the difficult tasks will be annotated by crowd workers again (line 11-16). In the end, we gather all the annotations, correct and aggregate them to compute the consensus labels (line 17-24).


\begin{algorithm}[h!tbp]
\caption{MetaCrowd: Crowdsourcing with \textbf{Meta}-workers}
\label{alg1}
\LinesNumbered
\KwIn{Project $\mathcal D$ ($N$ tasks of $n$ classes), related datasets $\mathcal{D}_{train}$, meta learning algorithm ($n$-way $k$-shot), meta-workers $\mathcal F$ (size $W_m$), divergence threshold $\theta$, crowd workers $\mathcal{W}^c$}
\KwOut{crowdsourcing consensus labels $\mathbf y$}
Cluster $\mathcal D$ into $n$ clusters\;
Query from $\mathcal{W}^c$ to obtain $\mathcal{D}_s$\;
{Use $\mathcal{D}_s$ and annotations to model $\mathcal{W}^c$\;}
$\mathcal{D}_q = \mathcal{D} - \mathcal{D}_s$, $\mathcal{D}_{test} = \mathcal{D}_s \cup \mathcal{D}_q$\;
Meta-workers $\mathcal{W}^m = \varnothing$\;
\For{$f$ in $\mathcal F$}{
    Use $f$ on $\mathcal{D}_{train}$ to extract meta-knowledge $\phi_f$\;
    Use $f$, $\phi_f$, and $\mathcal{D}_s$ to generate meta-worker $w^m_f$\;
    $\mathcal{W}^m = \mathcal{W}^m \cup \{w^m_f\}$\;
}
\For{$x_i$ in $\mathcal{D}_q$}{
    Use $\mathcal{W}^m$ to obtain annotations $\{\mathbf{A}^m_i\}_{m=1}^{W_m}$\;
    \If{$JS(\{\mathbf{A}^m_i\}_{m=1}^{W_m}) > \theta$}{
        Assign extra crowd workers to get $\{\mathbf{A}^c_i\}_{c=1}^{W_c}$\;
    }
}
Complete $\{\mathbf{A}^c\}_{c=1}^{W_c}$ with $\mathbf{a}_{pad} = [-1,-1, \cdots, -1]$\;
\While{not converged}{
    Correct crowd annotations using Eq. (\ref{mu})\;
    Correct meta annotations using Eq. (\ref{DS})\;
    Combine $\{\mathbf{A}^m\}_{m=1}^{W_m}\ \&\ \{\mathbf{A}^c\}_{c=1}^{W_c}$ to estimate consensus labels using Eq. (\ref{e-step})\;
    Update crowd workers' model and meta-workers' model using Eq. (\ref{m-step})\;
}
Return consensus labels $\mathbf y$\;
\end{algorithm}

\section{Experiments}
\subsection{Experimental Setup}
\textbf{Datasets}: We verified the effectiveness of our proposed method \textbf{MetaCrowd} on three real image datasets, \textbf{Mini-Imagenet} \cite{vinyals2016matching}, \textbf{256\_Object\_Categories} \cite{griffin2007caltech}, and \textbf{CUB\_200\_2011} \cite{wah2011caltech}. Each dataset has multiple subclasses and we treat each subclass in a dataset as a dependent task (or a small dataset). Following the dataset division principle recommended by Mini-Imagenet, we divide it into three parts: train, val and test, the category ratio is $64:16:20$, the other two datasets are also processed in a similar way. The statistics of these benchmark datasets are given in Table \ref{table:datasets}. We deem all the data in the `train' portion as a meta-training set and the data in the `val' portion as the validation set, and we randomly select $n$ categories from the `test' set to form an $n$-way meta-test set.

\begin{table}[h!tbp]
\caption{Statistics of the datasets.}
\begin{center}
\label{table:datasets}
\begin{tabular}{c | c c c}
\hline
\textbf{Dataset}        & $\mathbf{\mathcal{D}_{test}}$  & $\mathbf{\mathcal{D}_{train}}$     & \textbf{Image Num}   \\ \hline
{Mini-Imagenet}         & 20 class                       &  64+16 class                       & 600 * 100                       \\ \hline
{256\_Object}           & 40 class                       & 128+32 class                       & 90 * 200                        \\ \hline
{CUB\_200\_2011}        & 40 class                       & 128+32 class                       & 60 * 200                        \\ \hline
\end{tabular}
\end{center}
\end{table}

\textbf{Crowd Workers}: Following the canonical worker setting \cite{kazai2011worker}, we simulate four types of workers (spammer, random, normal, and expert), with different capacity (accuracy) ranges and proportions as shown in Table \ref{Worker}. We set up three different proportions of workers to study the influence of low reliability workers, normal reliability workers and high reliability workers, the weighted average capacity is 0.535, 0.600 and 0.650. We generate 30 crowd workers for Mini-Imagenet, and 10 crowd workers for 256\_Object\_Categories and CUB\_200\_2011 following the setup in Table \ref{Worker}, to initialize our $n$-way $k$-shot dataset and to provide additional annotations for the tasks when meta-workers disagree.

\begin{table}[h!tbp]
\caption{Crowd worker setup of proportion and capacity ranges. We simulated three groups of crowds with different worker type proportions whose comprehensive abilities ascended in turn. The second one is the typical setup we recommend.}
\begin{center}
\label{Worker}
\begin{tabular}{c|cc|ccc}
\hline
\textbf{Worker type} & \textbf{Floor} & \textbf{Ceiling} & \multicolumn{3}{c}{\textbf{Proportions}}                      \\ \hline
spammer              & 0.10           & 0.25             & \multicolumn{1}{c|}{10\%} & \multicolumn{1}{c|}{\textbf{10\%}} & 10\%  \\ \hline
random               & 0.25           & 0.50             & \multicolumn{1}{c|}{20\%} & \multicolumn{1}{c|}{\textbf{10\%}} & 10\%  \\ \hline
normal               & 0.50           & 0.80             & \multicolumn{1}{c|}{60\%} & \multicolumn{1}{c|}{\textbf{70\%}} & 50\%  \\ \hline
expert               & 0.80           & 1.00             & \multicolumn{1}{c|}{10\%} & \multicolumn{1}{c|}{\textbf{10\%}} & 30\%  \\ \hline
\end{tabular}
\end{center}
\end{table}

We compare our MetaCrowd against five related and representative methods. \\
(i) \textbf{Reqall} (Request and allocate) \cite{li2016crowdsourcing} is a typical solution for budget saving. Reqall dynamically determines the amount of annotation required for a given task. For each task, it stops further annotating if the weighted ratio between two classes’ vote counts reaching a preset threshold or the maximum number. Reqall assumes the workers' ability are known and mainly focus on binary task. We follow the advice in the paper to convert multi-class problems into binary ones. We fix its quality requirement as 3, consistent with MetaCrowd. \\
{(ii) \textbf{QASCA} \cite{zheng2015qasca} is a classical task assignment solution, it estimates the quality improvement if a worker is assigned with a set of tasks (from a pool of $N$ tasks), and then selects the optimal set which results in the highest expected quality improvement. We set the budget to an average of 3 annotations per task (similar to Reqall).} \\
(iii) \textbf{Active} (Active crowdsourcing) \cite{fang2014active} takes into account budget saving and worker/task selection for crowdsourcing. Active combines task domain (meta-test) and source domain (meta-training) data using sparse coding to get a more concise high-level representation of task, and then uses a probabilistic graphical model to model both workers and tasks, then uses active learning to select the right worker for right task. \\
(iv) \textbf{AVNC} (Adaptive voting noise correction) \cite{zhang2017improving} tries to identify and eliminate potential noisy annotations, then uses the remaining clean instances to train a set of classifiers to predict and correct the noisy annotations before truth inference. We use MV and WMV as its consensus algorithms and set the budget to an average of 3 annotations per task (similar to Reqall and QASCA)\\
(v) \textbf{ST} (Self-training method) \cite{xie2020self} first trains the annotator with labeled instances in the pool $\mathcal{D}_s$. The annotator then finishes the tasks in $\mathcal{D}_q$, picks out the instance with the highest confidence label and merges it into $\mathcal{D}_s$. The above steps are repeated until all tasks are labeled. \\
(vi) \textbf{MetaCrowd} and its variants adopt three meta-workers trained by three types of meta algorithms (MAML, MN, and RN) under the $5$-way $5$-shot setting. In our experiments, we consider two variants for the ablation study. \textbf{MetaCrowd-OC} follows the canonical crowdsourcing principle and uses only crowd workers. All the tasks are annotated by the three crowd workers, and we deem its accuracy and budget as the baseline performance. \textbf{MetaCrowd-OM} uses only meta-workers to annotate $\mathcal{D}_q$, even when they disagree. The remaining settings of the variants are the same as \textbf{MetaCrowd}.

For the other parameters not mentioned in the above comparison methods, we have adopted the recommended parameter settings in their original paper.


\begin{table*}[!htbp]
\caption{Accuracy of compared methods under three different crowd worker settings. The 1st/2nd columns of AVNC and MetaCrowd use weighted/unweighted majority vote as the consensus algorithm. Note that ST and MetaCrowd-OM have no crowd worker involved, so their accuracy does not change.}
\begin{center}

\subtable[Accuracy of compared methods with the crowd workers' weighted average capacity as 0.535 (more low-quality workers).]{
\begin{tabular}{c|cc|c|ccc|cccccc}
\hline
\textbf{Data}       &\textbf{Reqall} &\textbf{QASCA} &\textbf{Active} &\multicolumn{2}{c}{\textbf{AVNC}} &\textbf{ST} &\multicolumn{2}{c}{\textbf{MetaCrowd-OC}} &\multicolumn{2}{c}{\textbf{MetaCrowd-OM}} &\multicolumn{2}{c}{\textbf{MetaCrowd}} \\ \hline
\textbf{Mini-Imagenet}      & 0.778    & 0.781          & 0.733        & 0.768           & 0.808           & 0.307         & 0.624             & 0.763            & 0.698             & 0.764            & 0.748           & \textbf{0.825}           \\
\textbf{256\_Object}        & 0.785    & 0.796          & 0.740        & 0.773           & 0.812           & 0.315         & 0.634             & 0.771            & 0.679             & 0.757            & 0.752           & \textbf{0.828}           \\
\textbf{CUB\_200\_2011}     & 0.779    & 0.792          & 0.731        & 0.756           & 0.800           & 0.293         & 0.622             & 0.760            & 0.703             & 0.773            & 0.732           & \textbf{0.817}           \\ \hline
\end{tabular}
\label{ACC:a}
}

\subtable[Accuracy of compared methods with the crowd workers' weighted average capacity as 0.600.]{
\begin{tabular}{c|cc|c|ccc|cccccc}
\hline
\textbf{Data}       &\textbf{Reqall} &\textbf{QASCA} &\textbf{Active} &\multicolumn{2}{c}{\textbf{AVNC}} &\textbf{ST} &\multicolumn{2}{c}{\textbf{MetaCrowd-OC}} &\multicolumn{2}{c}{\textbf{MetaCrowd-OM}}  &\multicolumn{2}{c}{\textbf{MetaCrowd}} \\ \hline
\textbf{Mini-Imagenet}      & 0.821     & 0.824         & 0.775           & 0.792           & 0.833           & 0.307       & 0.672             & 0.806            & 0.698             & 0.764            & 0.767           & \textbf{0.840}           \\
\textbf{256\_Object}        & 0.828     & 0.827         & 0.777           & 0.807           & 0.838           & 0.315       & 0.664             & 0.792            & 0.679             & 0.757            & 0.778           & \textbf{0.855}           \\
\textbf{CUB\_200\_2011}     & 0.812     & 0.819         & 0.766           & 0.784           & 0.821           & 0.293       & 0.681             & 0.811            & 0.703             & 0.773            & 0.749           & \textbf{0.836}           \\ \hline
\end{tabular}
\label{ACC:b}
}

\subtable[Accuracy of compared methods with the crowd workers' weighted average capacity as 0.650 (more high-quality workers).]{
\begin{tabular}{c|cc|c|ccc|cccccc}
\hline
\textbf{Data}       &\textbf{Reqall} &\textbf{QASCA} &\textbf{Active} &\multicolumn{2}{c}{\textbf{AVNC}} &\textbf{ST} &\multicolumn{2}{c}{\textbf{MetaCrowd-OC}} &\multicolumn{2}{c}{\textbf{MetaCrowd-OM}} &\multicolumn{2}{c}{\textbf{MetaCrowd}} \\ \hline
\textbf{Mini-Imagenet}      & 0.869    & 0.862          & 0.813           & 0.842           & 0.888           & 0.307       & 0.749             & 0.860            & 0.698             & 0.764            & 0.834           & \textbf{0.903}           \\
\textbf{256\_Object}        & 0.858    & 0.880          & 0.817           & 0.857           & 0.891           & 0.315       & 0.757             & 0.858            & 0.679             & 0.757            & 0.842           & \textbf{0.911}           \\
\textbf{CUB\_200\_2011}     & 0.852    & 0.873          & 0.821           & 0.840           & 0.877           & 0.293       & 0.740             & 0.853            & 0.703             & 0.773            & 0.828           & \textbf{0.907}           \\ \hline
\end{tabular}
\label{ACC:c}
}

\end{center}
\label{ACC}
\end{table*}

\subsection{Analysis of the Results}
Table \ref{ACC} gives the accuracy of the methods under comparison, grouped into four categories: dynamic task allocation (Reqall and QASCA), active learning (Active), machine self correction (AVNC and ST), and meta learning (MetaCrowd-OC, MetaCrowd-OM, and MetaCrowd). Particularly, AVNC and MetaCrowd adopt majority vote and weighted majority vote to compute consensus labels, while other methods adopts their own consensus solutions. We have several important observations. \\
(i) \textbf{MetaCrowd vs. Self-training:} {ST} uses supervised self-training to gradually annotate the tasks, it has the lowest accuracy among all the compared methods. This is because the lack of meta-knowledge (labeled training data) makes traditional supervised methods unfeasible under the setting of few-shot learning. When the number of labeled instances is small, ST cannot train an effective model, so the quality of pseudo-labels is not high, and the influence of the error will continue to expand, eventually leading to the failure of the self-training classifier. The other crowdsourcing solutions obtain a much higher accuracy by modelling tasks and workers, and MetaCrowd achieves the best performance through both meta learning and ensemble learning. This shows that the extraction of meta-knowledge from relevant domains is crucial for the few-shot learning process. \\
(ii) \textbf{MetaCrowd vs. Active:} Both {MetaCrowd} and {Active} try to reduce the number of annotations to save the budget. By modeling workers and tasks, {Active} assigns only the single most appropriate worker to the task to save the budget and to improve the quality. In contrast, MetaCrowd leverages meta-knowledge and initial labels from crowd workers to automatically annotate a large portion of simple tasks, and invites crowd workers to annotate a small number of difficult tasks. Thus, both quality and budget saving can be achieved. {MetaCrowd-OM} and {MetaCrowd} both achieve a significantly higher accuracy than {Active}. This is because crowd workers have diverse preferences and the adopted active learning strategy of Active cannot reliably model workers due to the limited data. By obtaining additional annotations for the most uncertain tasks and meta annotations for all the tasks, {MetaCrowd} achieves an accuracy improvement of 8\%. Even with a large number of annotations from plain crowd workers, {MetaCrowd-OC} still loses to {MetaCrowd}, which proves the rationality of empowering crowdsourcing with meta learning. \\
(iii) \textbf{MetaCrowd vs. Reqall:} Both {MetaCrowd} and {Reqall} dynamically determine the number of annotations required for a task based on the annotating results; thus they can trade-off budget with quality. {Reqall} assumes that workers' abilities are known and mainly focuses on binary tasks (we follow the recommended approach to adjust it to the multi-class case). With no more than three annotations per task on average ({MetaCrowd-OC} baseline setting), {MetaCrowd} beats {Reqall} both in quality and budget, because {MetaCrowd} can leverage the classifiers to do most of the simple tasks, and save the budget to focus on the difficult tasks. These results show the effectiveness of our human-machine hybrid approach. \\
{(iv) \textbf{MetaCrowd vs. QASCA:} QASCA and Reqall both decide the next task assignment based on the current annotation results. QASCA does not assume that the workers' abilities are known, but derives them based on the EM algorithm, so its actual performance is much better than Reqall. However, QASCA is still beaten by MetaCrowd. This is because QASCA is plagued by the cold start problem. At the beginning, it can only treat all workers as saints, and is more affected by low-quality workers. In contrast, MetaCrowd has a relatively accurate understanding of workers at the beginning, owing to the meta-test set construction process and the repel of low-quality workers for difficult tasks.} \\
(v) \textbf{Modeling vs. non-modeling of workers:} In crowdsourcing, we often need to model tasks and/or workers to better perform the tasks and compute the consensus labels of the tasks. Comparing the second column (weighted majority vote) of {AVNC}, {MetaCrowd-OC}, {MetaCrowd-OM} and {MetaCrowd} with their respective non-model version (first column, majority vote), we can draw the conclusion that, by modeling workers, we can better compute the consensus labels of tasks from their annotations. This advantage is due to two factors: we introduce a worker model to separately account for crowd workers and meta-workers; and we model the difficulty of tasks using divergence, and pay more attention to the difficult ones. \\
(vi) \textbf{Robustness to different situations:} We treat the results in Table \ref{ACC:b} as the baseline, in Table \ref{ACC:a} there are more noisy workers and while in Table \ref{ACC:c} there are more experts. By comparing the results in Table \ref{ACC:a} and Table \ref{ACC:b}, we can see that although the average quality of crowd workers drops by $6.5\%$, the accuracy of MetaCrowd reduces by less than $2\%$, while other methods have a larger reduce $3\% \sim 6\%$. This can be explained by two reasons. First, MetaCrowd utilizes sufficient data to model workers during building the Meta-test set stage, so potential low-quality crowd workers can be identified; second, most of the simple task annotations are provided by ordinary but reliable meta-workers, and we also consult crowd workers with fair quality for difficult tasks. Therefore, MetaCrowd can reduce the impact of low-quality workers and obtain more robust results. In even less common situations with many experts (Table \ref{ACC:c}), MetaCrowd also achieves the best performance. This confirms that our MetaCrowd is suitable for a variety of worker compositions, especially when the capacity of workers is not so good.

\subsection{Budget Saving}
We use quantitative analysis and simulation results to illustrate the advantage of MetaCrowd for budget saving in terms of the number of used annotations. For this quantitative analysis, we adopt the typical assumption that the expense of a single annotation is uniformly fixed. We separately estimate the budget of {MetaCrowd-OC}, {Reqall}, {QASCA}, {Active} and {MetaCrowd}. {AVNC} adopts the same workflow as {MetaCrowd-OC}, except for noise correction operation, so its budget is the same as MetaCrowd-OC. On the other hand, ST and MetaCrowd-OM basically has no crowd workers involved, so they are not considered here.

\subsubsection{Quantitative Analysis}
The number of annotations used by some methods can be estimated in advance, we first calculate them theoretically. \\
\textbf{MetaCrowd-OC} asks $W_m$ crowd workers to annotate each task and consumes a total of $(N W_m)$ annotations. \\
\textbf{MetaCrowd} needs some initial annotations $\mathcal{D}_s$ to kick off the training of meta-workers. The total number of annotations needed in {MetaCrowd} is $\gamma n k W_c + (N - \gamma n k) \beta W_m$, where $n k$ is the size of $\mathcal{D}_s$, $W_c$ is the number of crowd workers for repeated annotations, $\gamma > 1$ is the margin amplification factor to ensure that $\mathcal{D}_s$ can be formed, and $\beta < 1$ is the ratio (determined by the divergence threshold $\theta$) of instances that need additional annotations (the other symbols can be found in Table \ref{marks}). Usually $N \gg (n k, W_c) > (s, \gamma) > 1 > \beta$, if $N$ is big enough, then the number of annotations can be simplified as $\beta N W_m$. \\
\textbf{Active} needs about $\frac{1}{3} N$ annotated tasks to build a stable model of workers and tasks before applying active learning; each of the remaining tasks only needs the single most suitable crowd worker. We let $W_m$ crowd workers annotate $\frac{1}{3} N$ tasks, so the total number of annotations for {Active} is $\frac{1}{3}N W_m + \frac{2}{3}N = \frac{W_m+2}{3} N$. \\
\textbf{QASCA} has set the total budget to an average of 3 annotations per task, so it also consumes a total of $3N$ annotations. \\
\textbf{Reqall} depends on the quality requirement, task difficulty, and workers' ability to determine the number of consumed annotations, so the required number of annotations can not be explicitly quantified. In the multi-class case, Reqall considers the two classes with the most votes and thus wastes the budget to some extent. \\

In summary, the total number of annotations needed in {MetaCrowd-OC}, {Active}, {MetaCrowd} and {QASCA} are about $N W_m$, $\frac{W_m+2}{3} N$, $\beta N W_m$ and $3N$, respectively.

\subsubsection{Simulation Results}
Following the experimental settings in the previous subsection, we count the number of consumed crowd worker annotations on Mini-Imagenet ($N = 3000$, $W_m = 3$) for example. \\
Both \textbf{MetaCrowd-OC} and \textbf{QASCA} require $3000 \times 3 = 9000$ annotations; \\
\textbf{MetaCrowd} ($n=k=5,\ W_c=30,\ \beta \approx 1/3,\ \gamma \approx 1.34$) costs $1.34 \times 5 \times 5 \times 30 + (3000 - 1.34 \times 5 \times 5) \times \frac{1}{3} \times 3 = 3971$ annotations. \\
\textbf{Active} asks for about $3000 \times \frac{1}{3} \times 3 = 3000$ annotations to build the model, and the other tasks consumes about $3000 \times \frac{2}{3} \times 1 = 2000$ annotations, for a total number of about $5000$ annotations. \\
\textbf{Reqall}, with the requirement that the budget should be no more than three annotations per task on average, achieves a competitive quality and consumes about 2.8 annotations per task, amount to a total of $2.8 \times 3000 = 8400$ annotations; \\

\begin{table*}[!htbp]
\caption{Number of annotations used and accuracy of {MetaCrowd-OC}, {Reqall}, {Active} and {MetaCrowd}. Number in column `Budget' means the number of used annotations and number in column `Quality' means the consensus label accuracy. We use the typical crowd worker proportion in Table \ref{Worker}.}
\begin{center}
\begin{tabular}{c|c|cc|cc|cc|cc|cc}
\hline
\multirow{2}{*}{\textbf{Data}} & \multirow{2}{*}{\textbf{Task}} & \multicolumn{2}{c|}{\textbf{Reqall}} & \multicolumn{2}{c|}{\textbf{QASCA}} & \multicolumn{2}{c|}{\textbf{Active}} & \multicolumn{2}{c|}{\textbf{MetaCrowd-OC}} & \multicolumn{2}{c}{\textbf{MetaCrowd}} \\ \cline{3-12}
                               &                             & \textbf{Budget}      & \textbf{Quality}     & \textbf{Budget}      & \textbf{Quality}  & \textbf{Budget}      & \textbf{Quality}   & \textbf{Budget}      & \textbf{Quality}      & \textbf{Budget}    & \textbf{Quality}    \\ \hline
\textbf{Mini-Imagenet}              & 3000                        & 8424              & 0.821       &9000     &  0.824   & 5000              & 0.775            & 9000              & 0.806             & \textbf{3971}   & \textbf{0.840}  \\
\textbf{256\_Object}                & 450                         & 1256              & 0.828       &1350    &   0.827  & 750               & 0.777            & 1350              & 0.792             & \textbf{654}    & \textbf{0.855}  \\
\textbf{CUB\_200\_2011}             & 300                         & 837               & 0.812       &900    &   0.819  & \textbf{500}      & 0.766            & 900               & 0.811             & 530             & \textbf{0.836}  \\ \hline
\end{tabular}
\label{budget}
\end{center}
\end{table*}

The total number of annotations and accuracy of the methods on three datasets are listed in Table \ref{budget}. {MetaCrowd} always outperforms {Active}, {Reqall}, and {MetaCrowd-OC} in terms of budget and quality, and its budget saving advantage becomes more prominent as the increase of crowdsourcing project size $N$. {MetaCrowd} loses to {Active} in terms of budget on {CUB\_200\_2011}, due to the small size of this dataset. In fact, {MetaCrowd} is not suitable for crowdsourcing tasks with a relatively small size, or with an extremely large label space, in which repeated annotations of $\mathcal{D}_s$ for meta-learning consume a large portion of the budget.


\subsection{Parameter Analysis}
\subsubsection{Parameters in Meta Learning}
We study the impact of some preset parameters of MetaCrowd, namely $n$ and $k$ for meta learning algorithm. Generally, the values of $n$ and $k$ are determined by the given problem. Here we simulate the influence of $n$ and $k$ on the meta learning algorithm using Mini-Imagenet with MAML algorithm, other datasets and algorithm combinations lead to similar conclusions. We vary $k$ or $n$ from 1 to 10 while keeping the other fixed as in the previous experiments.

\begin{figure}[h!tbp]
\centerline{\includegraphics[width=9cm]{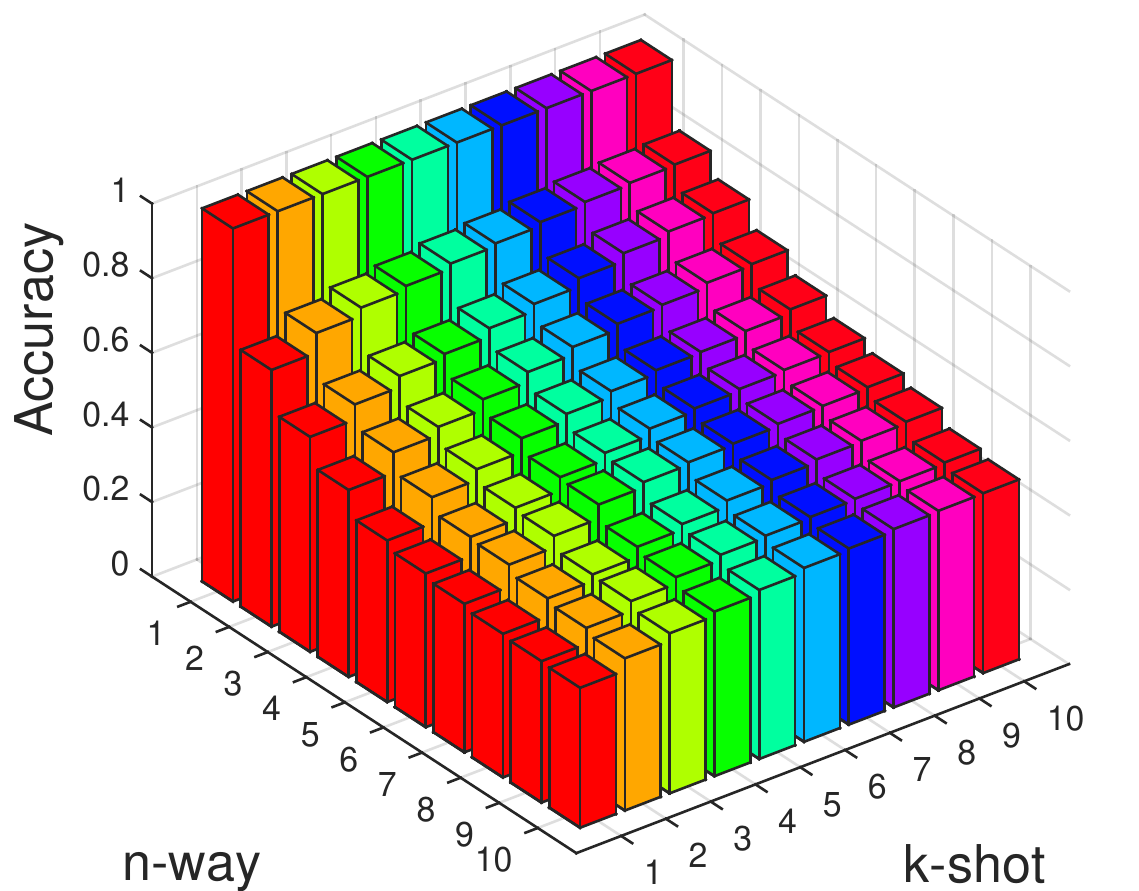}}
\caption{Accuracy vs. the number of class labels $n$ and the number of training instances $k$ per label. Note that as $n$ increase, the accuracy of random guessing closes to $1/n$.}
\label{fig:nk}
\end{figure}

We can see from Fig. \ref{fig:nk} that the accuracy increases as the number of tasks increases, but the increment gradually slows down, which is consistent with our intuition that more annotated tasks facilitate the training of more credible meta-workers, and hence improve the quality. On the other hand, as the number of classes increases, the accuracy gradually decreases, but it is always much higher than that of random guess, which suggests the effectiveness of the meta learning algorithm for a few-shot learning task. In crowdsourcing, $n$ is determined by the crowdsourcing project itself, and the only parameter we can choose is $k$. A larger $k$ gives a better meta-worker, but it is expensive to form a meta-training set $D_s$ with a large $k$; as such, $k=5$ is adopted as a trade-off in this paper.

\begin{figure*}[h!tbp]
    \centering
        \subfigure[MAML]{\includegraphics[width=6cm]{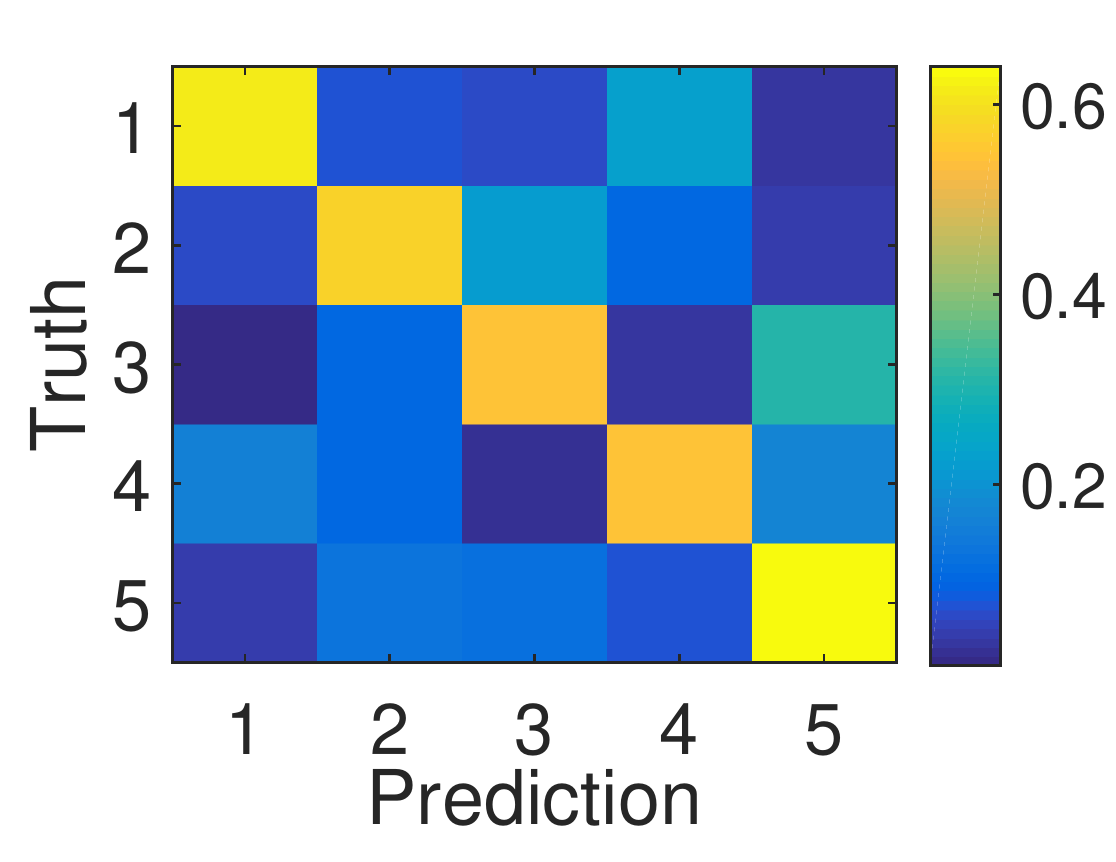}}
        \subfigure[MN]{\includegraphics[width=6cm]{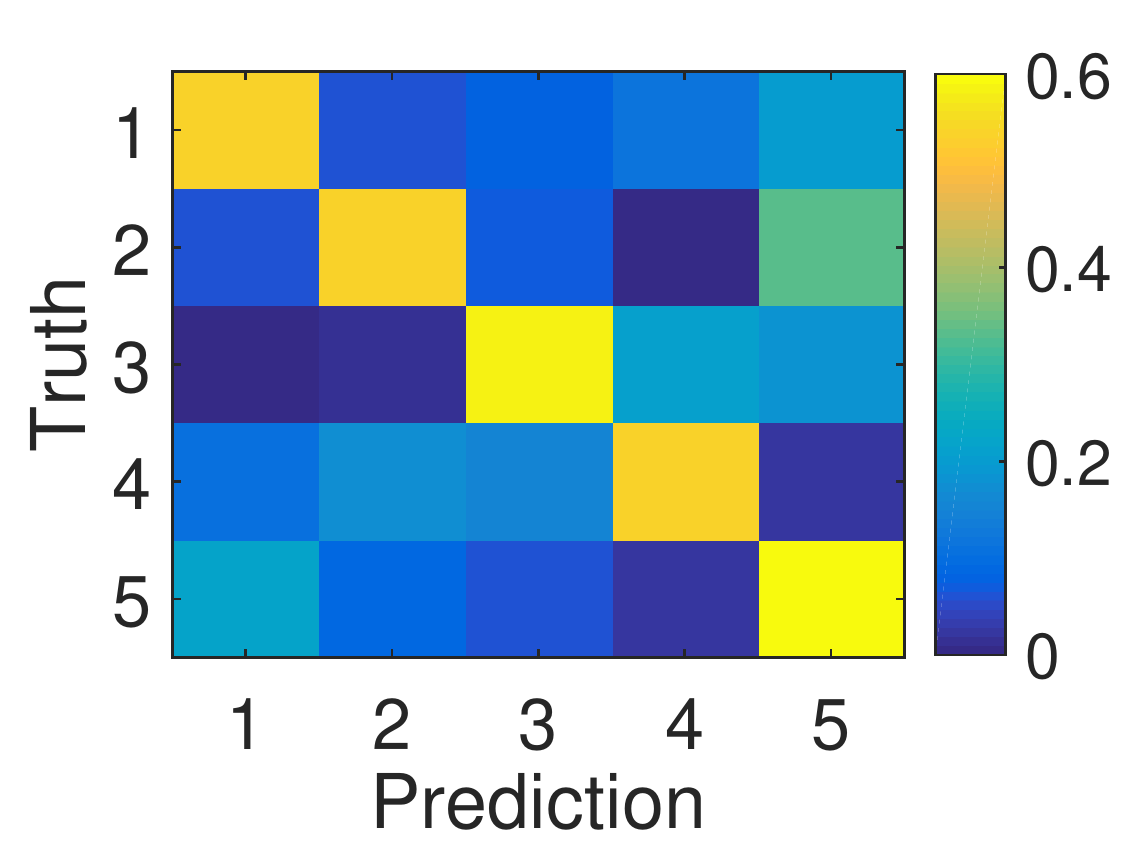}}
        \subfigure[RN]{\includegraphics[width=6cm]{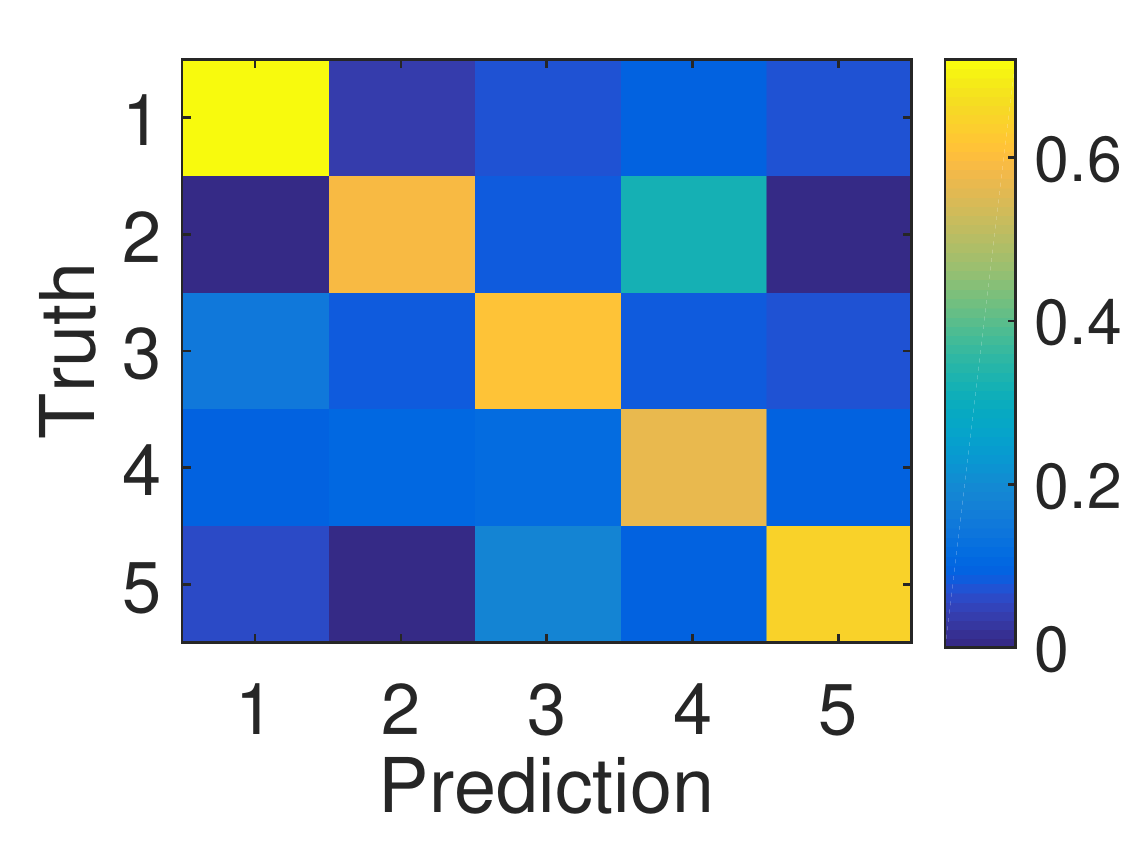}}
    \caption{Heat-maps of confusion matrices of three meta-workers (MAML, MN and RN). It can be seen that the accuracy of the meta-workers is around 0.6, and these meta-workers have different induction preferences, which are beneficial for ensemble learning.}
    \label{heat}
\end{figure*}

The heat-maps of the confusion matrices of our meta-workers under a 5-way 5-shot setting are shown in Fig. \ref{heat}. We find that the accuracy of all three meta-workers is about 0.6 (the values on the diagonal of each confusion matrix), which is in accordance with the average capacity of our normal crowd workers in Table \ref{Worker}. In addition, the meta-workers also manifest different preferences and are capable of doing different tasks.

\subsubsection{Parameters in Crowdsourcing}
Here we study the impact of the divergence threshold $\theta$ and the number of additional annotations $N_{add}$ for difficult task on the crowdsourcing quality and budget trading off. Generally speaking, a larger $\theta$ and $N_{add}$ will lead to a better crowdsourcing quality while consuming more budget as well, so how to set appropriate values for these two parameters to meet the quality and budget requirements as much as possible is critical.

\begin{figure*}[h!tbp]
    \centering
        \subfigure[Mini-Imagenet]{\includegraphics[width=6cm]{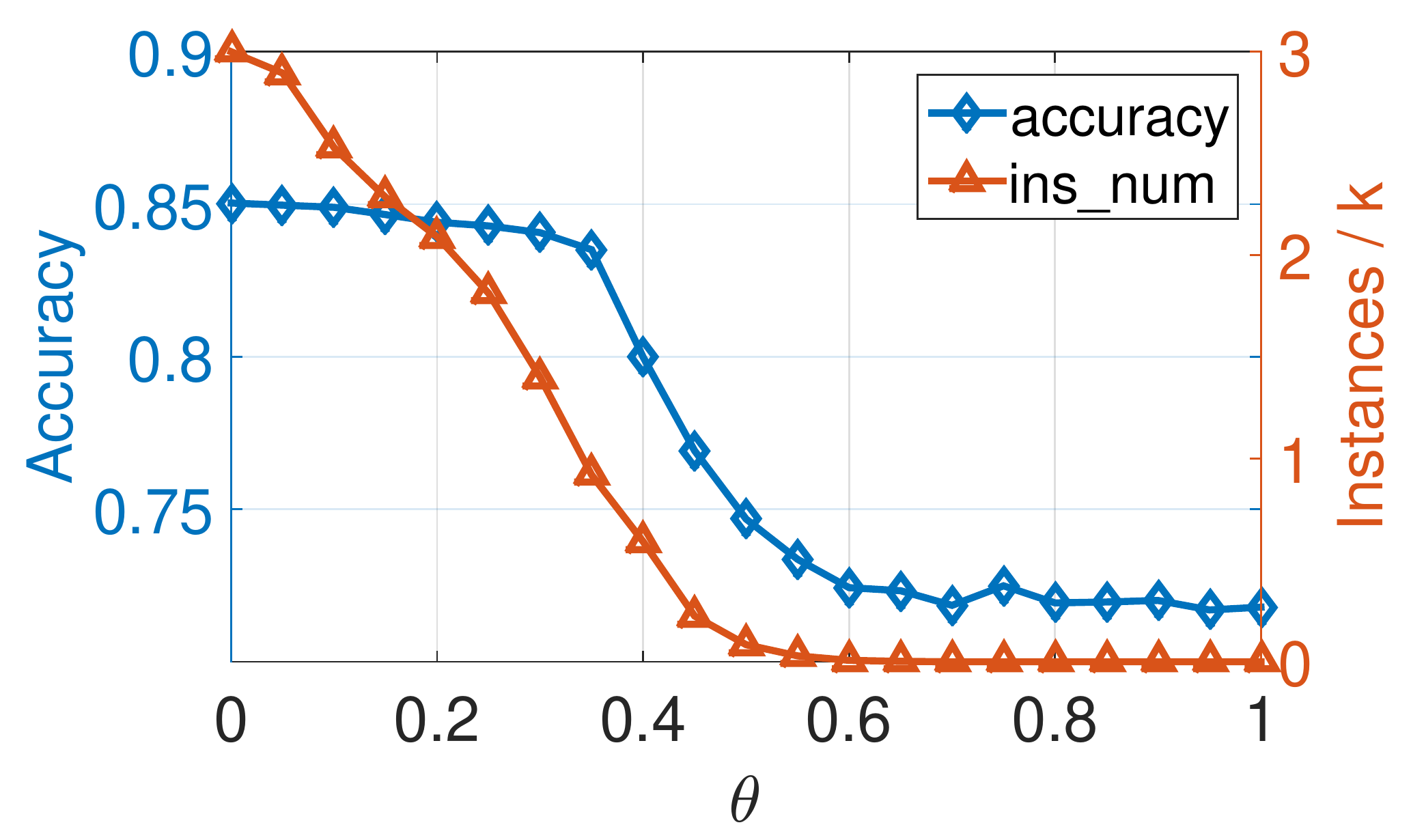}}
        \subfigure[256\_Object]{\includegraphics[width=6cm]{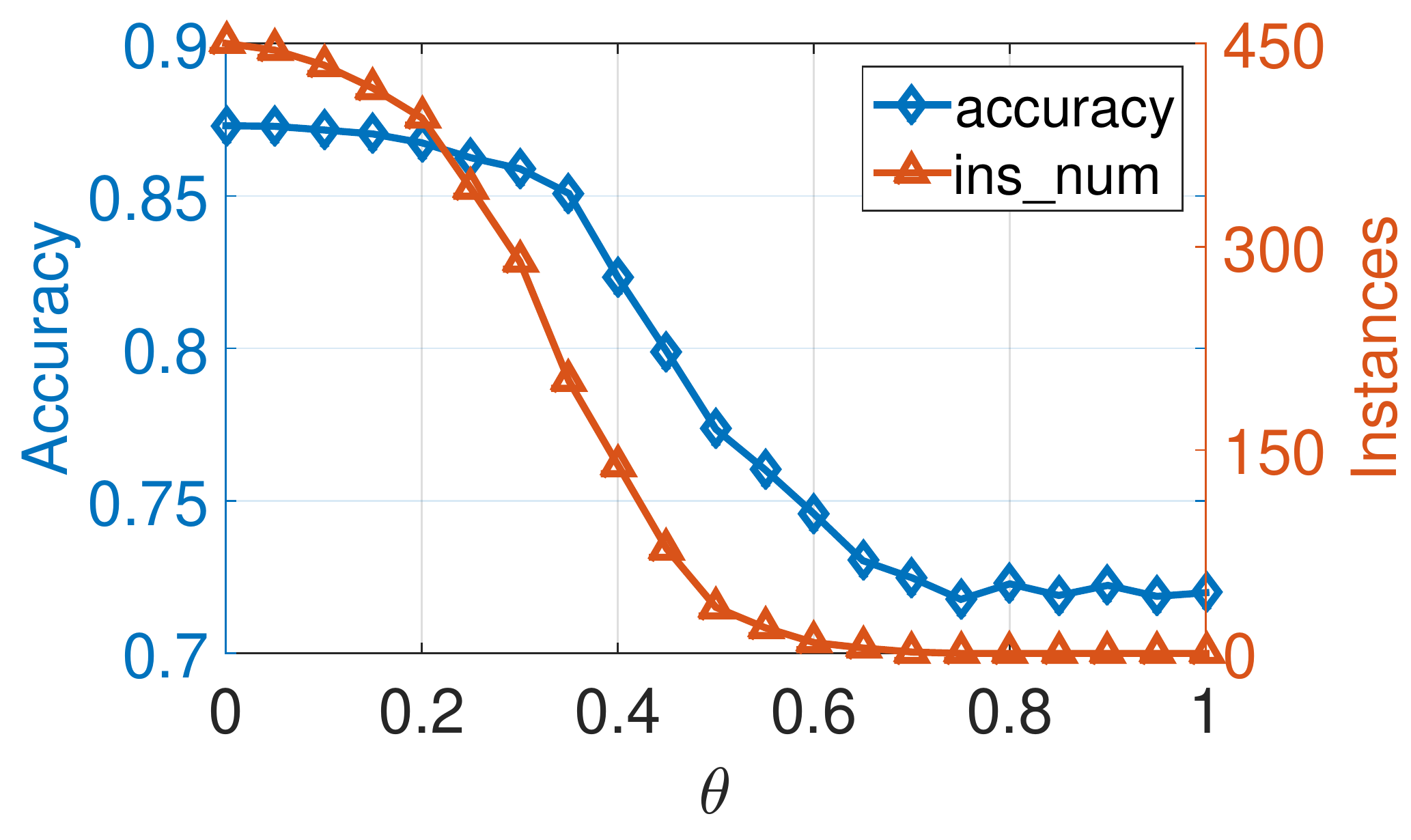}}
        \subfigure[CUB\_200\_2011]{\includegraphics[width=6cm]{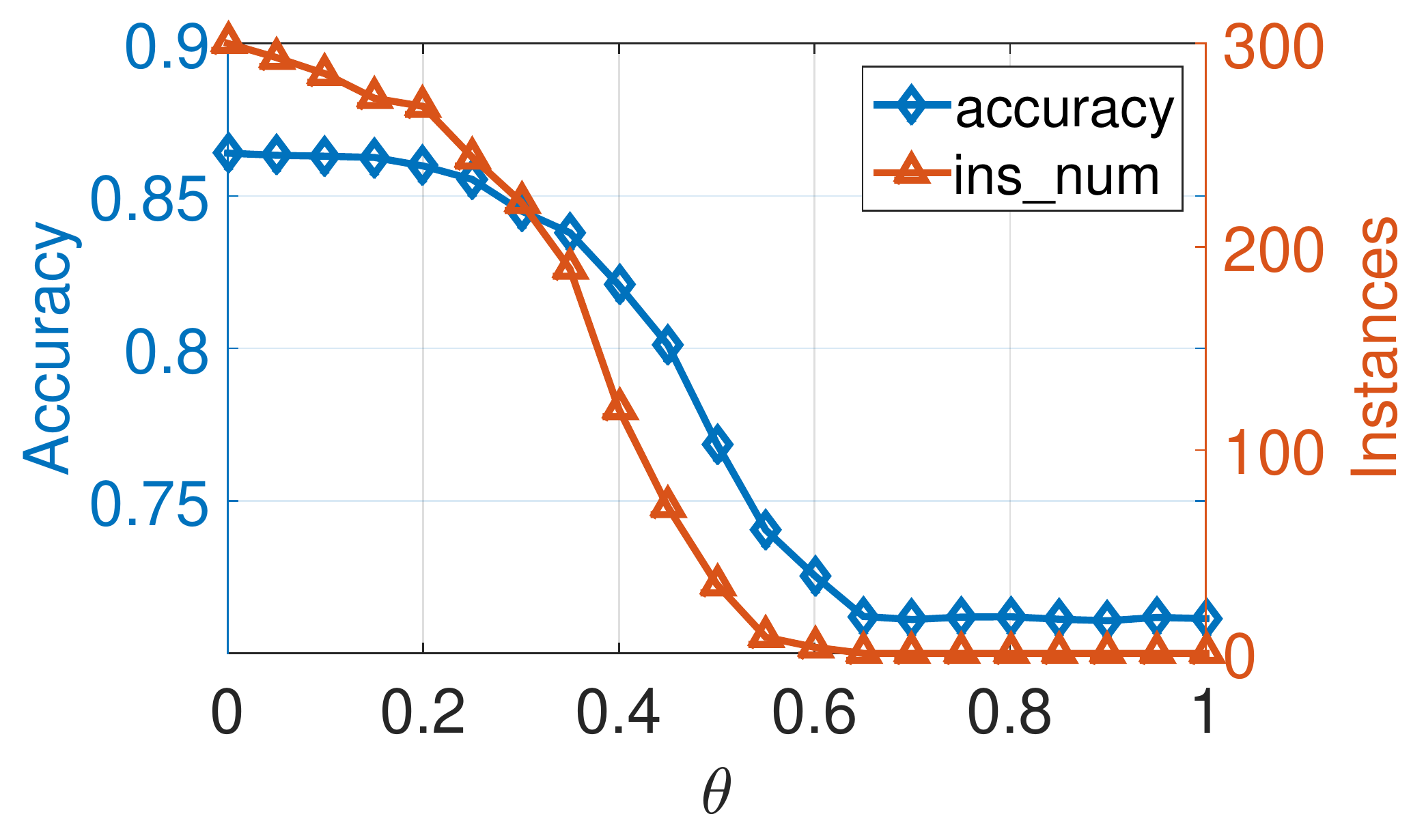}}
    \caption{Accuracy and the number of (\textit{kilo}) difficult tasks with the change of divergence threshold $\theta$ on three datasets. }
\label{fig:parameter}
\end{figure*}

The divergence quantification metric (see Eq. (\ref{mJS})) is within $(0, 1)$, so we change $\theta$ from 0 to 1 with an interval of 0.05. For each $\theta$ value, we assign $N_{add} = W_m$ workers to the task whose divergence degree exceeds $\theta$, and finally count the number of instances that received further crowd annotation. Fig. \ref{fig:parameter} gives the results under different input values of $\theta$.
We can see that as $\theta$ decreases from 1 to 0, the number of instances that need additional annotations gradually increases, and the aggregation accuracy also increases. The overall trend can be roughly divided into three stages according to the value of $\theta$: $[1, 0.6], [0.6, 0.3], [0.3, 0]$. In the first stage $\theta \in [1, 0.6]$, although there is a large variation range of $\theta$, the number of tasks needs to be manually annotated and the quality of the crowdsourcing project are almost unchanged, this is because the divergence of tasks is roughly distributed within the range of $[0, 0.6]$, so the influence of $\theta$ in the first stage is very limited. However in the second stage, the budget and quality of crowdsourcing project are very sensitive to $\theta$. With the decrease of $\theta$, both budget and quality increase significantly. In our experiment, setting the value of $\theta$ within [0.3, 0.6] is a reasonable choice. As to the third stage, the budget of the crowdsourcing project is still increasing rapidly with the change of $\theta$, but the quality keeps a relatively stable stage with little improvement. This can be explained as that the tasks with a small degree of divergence are usually relatively simple tasks, and their annotation results are agreed between meta-workers, and the further manual annotations will not significantly improve the quality. Based on these results, we adopt $\theta = 0.33$ for experiments, and for the trade off between quality and budget.

\begin{figure*}[h!tbp]
    \centering
        \subfigure[Mini-Imagenet]{\includegraphics[width=6cm]{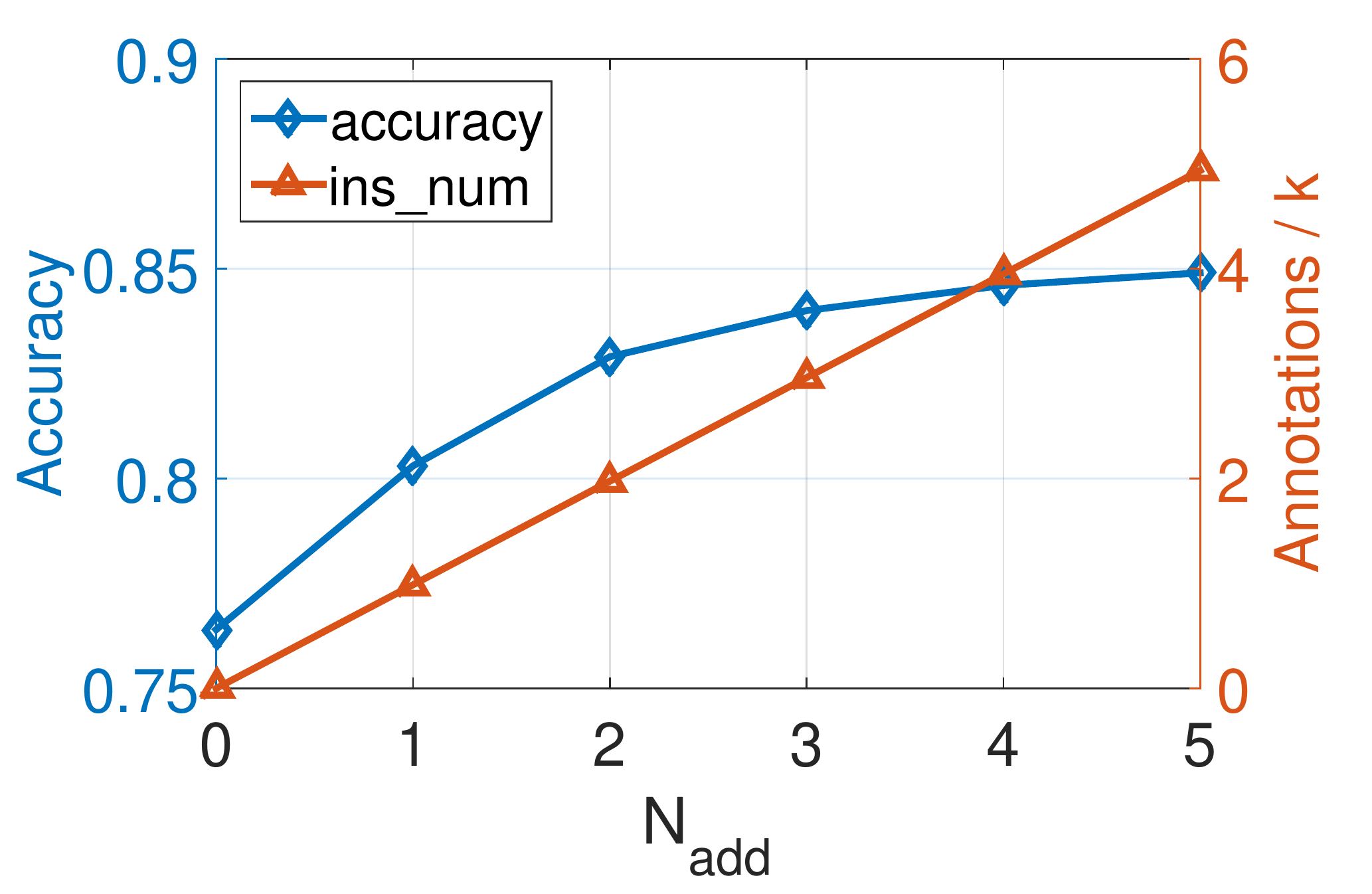}}
        \subfigure[256\_Object]{\includegraphics[width=6cm]{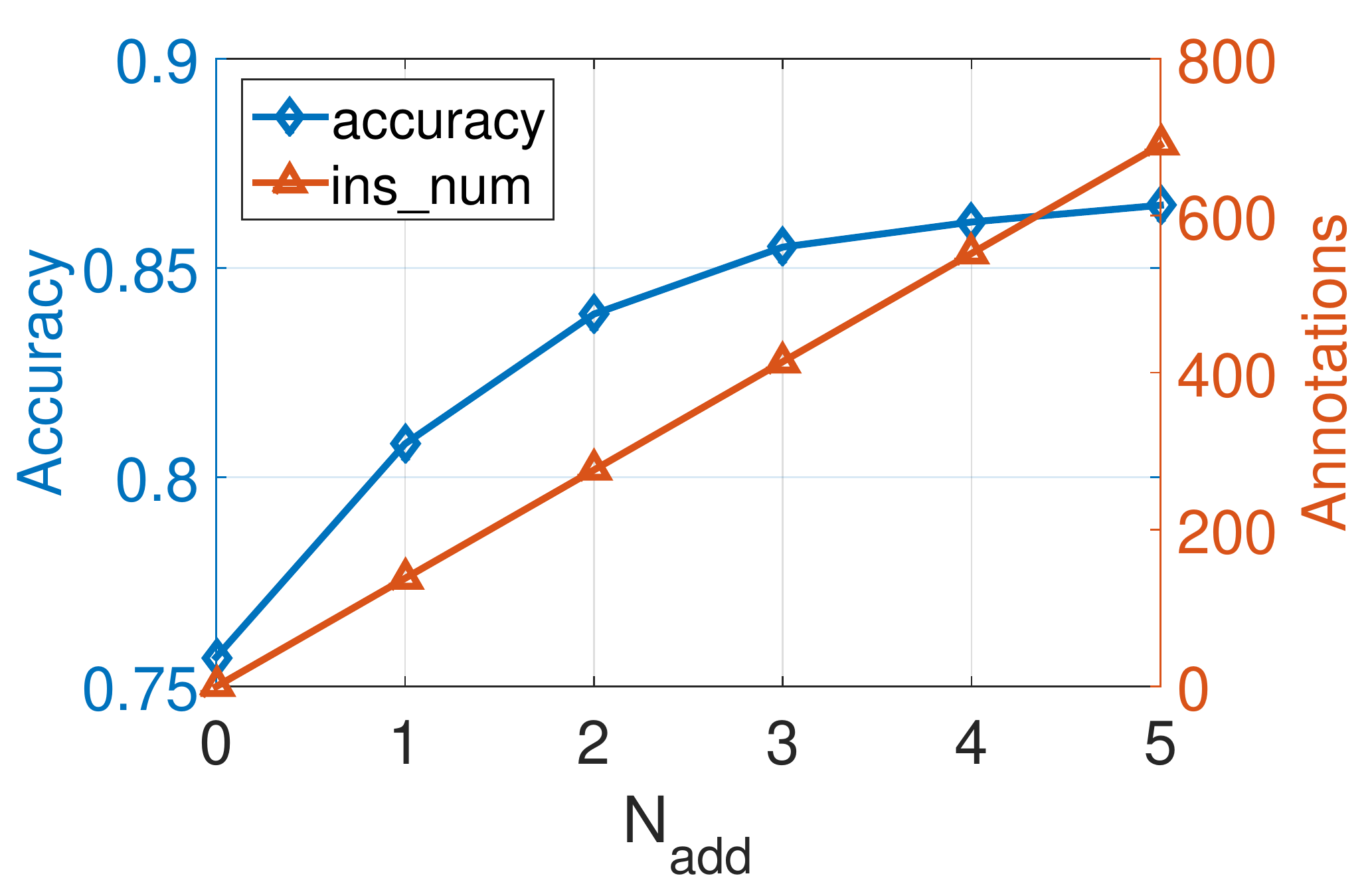}}
        \subfigure[CUB\_200\_2011]{\includegraphics[width=6cm]{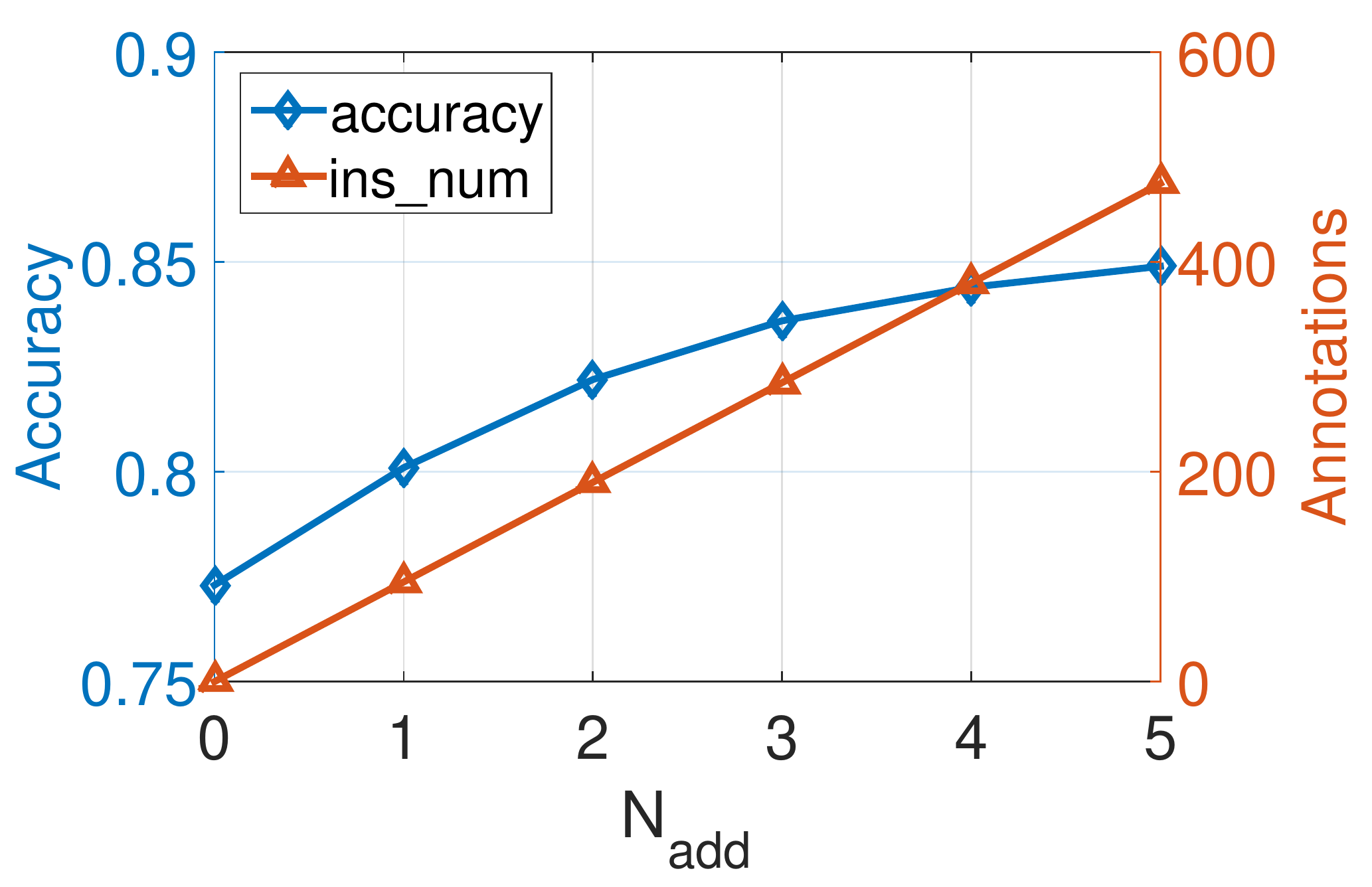}}
    \caption{Accuracy and the number of (\textit{kilo}) extra annotations with the change of $N_{add}$ annotations received for each difficult task on three datasets.}
\label{Nadd}
\end{figure*}

We also studied the impact of the number of additional annotations $N_{add}$ received for each difficult task on the quality and budget of crowdsourcing. The experimental results in Fig. \ref{Nadd} are in line with our intuition: when $N_{add}$ increases, the quality of crowdsourcing will gradually increase and becomes relatively stable afterwards. At the same time, the accompany budget linearly increases. If $N_{add} = 0$, MetaCrowd will degenerate into MetaCrowd-OM. Given that, we fix $N_{add}=3$, which is the same as the number of meta-workers $W_m$ in our experiment.

\section{Conclusion}
In this paper, we study how to leverage meta learning with crowdsourcing for budget saving and quality improving, and propose an approach called MetaCrowd (Crowdsourcing with Meta-Workers) that implements this idea. Our {MetaCrowd} approach uses meta learning to train capable meta-workers for crowdsourcing tasks and thus to save budgets. Meanwhile, it quantifies the divergence between meta-workers' annotations to model the difficulty of tasks, and collects additional annotations for difficult tasks from crowd workers to improve the quality. Experiments on benchmark datasets show that MetaCrowd is superior to the representative methods in terms of budget saving and crowdsourcing quality.

Our method has a better tolerance for the quality of crowdsourcing workers, but it has certain requirements for the types and characteristics of crowdsourcing tasks. One of the possible improvement of our work lies in how to obtain the initial labeled data set $D_s$ required for meta learning with less budget.

\bibliographystyle{IEEEtran}
\bibliography{MetaCrowd_Bib}

\begin{thebibliography}{10}
\providecommand{\url}[1]{#1}
\csname url@samestyle\endcsname
\providecommand{\newblock}{\relax}
\providecommand{\bibinfo}[2]{#2}
\providecommand{\BIBentrySTDinterwordspacing}{\spaceskip=0pt\relax}
\providecommand{\BIBentryALTinterwordstretchfactor}{4}
\providecommand{\BIBentryALTinterwordspacing}{\spaceskip=\fontdimen2\font plus
\BIBentryALTinterwordstretchfactor\fontdimen3\font minus
  \fontdimen4\font\relax}
\providecommand{\BIBforeignlanguage}[2]{{%
\expandafter\ifx\csname l@#1\endcsname\relax
\typeout{** WARNING: IEEEtran.bst: No hyphenation pattern has been}%
\typeout{** loaded for the language `#1'. Using the pattern for}%
\typeout{** the default language instead.}%
\else
\language=\csname l@#1\endcsname
\fi
#2}}
\providecommand{\BIBdecl}{\relax}
\BIBdecl

\bibitem{howe2006rise}
J.~Howe, ``The rise of crowdsourcing,'' \emph{Wired Magazine}, vol.~14, no.~6,
  pp. 1--4, 2006.

\bibitem{sheng2019machine}
V.~S. Sheng and J.~Zhang, ``Machine learning with crowdsourcing: A brief
  summary of the past research and future directions,'' in \emph{Proceedings of
  the AAAI Conference on Artificial Intelligence}, vol.~33, 2019, pp.
  9837--9843.

\bibitem{li2016crowdsourced}
G.~Li, J.~Wang, Y.~Zheng, and M.~J. Franklin, ``Crowdsourced data management: A
  survey,'' \emph{IEEE Transactions on Knowledge and Data Engineering},
  vol.~28, no.~9, pp. 2296--2319, 2016.

\bibitem{chittilappilly2016survey}
A.~I. Chittilappilly, L.~Chen, and S.~Amer-Yahia, ``A survey of general-purpose
  crowdsourcing techniques,'' \emph{IEEE Transactions on Knowledge and Data
  Engineering}, vol.~28, no.~9, pp. 2246--2266, 2016.

\bibitem{sheng2008get}
V.~S. Sheng, F.~Provost, and P.~G. Ipeirotis, ``Get another label? improving
  data quality and data mining using multiple, noisy labelers,'' in \emph{ACM
  SIGKDD International Conference on Knowledge Discovery and Data Mining},
  2008, pp. 614--622.

\bibitem{zheng2016docs}
Y.~Zheng, G.~Li, and R.~Cheng, ``Docs: a domain-aware crowdsourcing system
  using knowledge bases,'' \emph{VLDB Endowment}, vol.~10, no.~4, pp. 361--372,
  2016.

\bibitem{boim2012asking}
R.~Boim, O.~Greenshpan, T.~Milo, S.~Novgorodov, N.~Polyzotis, and W.-C. Tan,
  ``Asking the right questions in crowd data sourcing,'' in \emph{28th IEEE
  International Conference on Data Engineering}, 2012, pp. 1261--1264.

\bibitem{welinder2010multidimensional}
P.~Welinder, S.~Branson, P.~Perona, and S.~Belongie, ``The multidimensional
  wisdom of crowds,'' \emph{Advances in Neural Information Processing Systems},
  vol.~23, pp. 2424--2432, 2010.

\bibitem{zheng2015qasca}
Y.~Zheng, J.~Wang, G.~Li, R.~Cheng, and J.~Feng, ``Qasca: A quality-aware task
  assignment system for crowdsourcing applications,'' in \emph{ACM SIGMOD
  International Conference on Management of Data}, 2015, pp. 1031--1046.

\bibitem{li2016crowdsourcing}
Q.~Li, F.~Ma, J.~Gao, L.~Su, and C.~J. Quinn, ``Crowdsourcing high quality
  labels with a tight budget,'' in \emph{ACM International Conference on Web
  Search and Data Mining}, 2016, pp. 237--246.

\bibitem{tu2020attention}
J.~Tu, G.~Yu, J.~Wang, C.~Domeniconi, and X.~Zhang, ``Attention-aware answers
  of the crowd,'' in \emph{SIAM International Conference on Data Mining}, 2020,
  pp. 451--459.

\bibitem{vanschoren2018meta}
J.~Vanschoren, ``Meta-learning: A survey,'' \emph{arXiv preprint
  arXiv:1810.03548}, 2018.

\bibitem{sung2018learning}
F.~Sung, Y.~Yang, L.~Zhang, T.~Xiang, P.~H. Torr, and T.~M. Hospedales,
  ``Learning to compare: Relation network for few-shot learning,'' in
  \emph{IEEE Conference on Computer Vision and Pattern Recognition}, 2018, pp.
  1199--1208.

\bibitem{kazai2011worker}
G.~Kazai, J.~Kamps, and N.~Milic-Frayling, ``Worker types and personality
  traits in crowdsourcing relevance labels,'' in \emph{ACM International
  Conference on Information and Knowledge Management}, 2011, pp. 1941--1944.

\bibitem{liu2020FSLHierarchy}
L.~Liu, T.~Zhou, G.~Long, J.~Jiang, and C.~Zhang, ``Many-class few-shot
  learning on multi-granularity class hierarchy,'' \emph{IEEE Transactions on
  Knowledge and Data Engineering}, vol.~99, no.~1, pp. 1--14, 2020.

\bibitem{finn2017model}
C.~Finn, P.~Abbeel, and S.~Levine, ``Model-agnostic meta-learning for fast
  adaptation of deep networks,'' in \emph{International Conference on Machine
  Learning}, 2017, pp. 1126--1135.

\bibitem{santoro2016meta}
A.~Santoro, S.~Bartunov, M.~Botvinick, D.~Wierstra, and T.~Lillicrap,
  ``Meta-learning with memory-augmented neural networks,'' in
  \emph{International Conference on Machine Learning}, 2016, pp. 1842--1850.

\bibitem{munkhdalai2017meta}
T.~Munkhdalai and H.~Yu, ``Meta networks,'' in \emph{International Conference
  on Machine Learning}, 2017, pp. 2554--2563.

\bibitem{bromley1993signature}
J.~Bromley, J.~W. Bentz, L.~Bottou, I.~Guyon, Y.~LeCun, C.~Moore,
  E.~S{\"a}ckinger, and R.~Shah, ``Signature verification using a “siamese”
  time delay neural network,'' \emph{International Journal of Pattern
  Recognition and Artificial Intelligence}, vol.~7, no.~04, pp. 669--688, 1993.

\bibitem{snell2017prototypical}
J.~Snell, K.~Swersky, and R.~Zemel, ``Prototypical networks for few-shot
  learning,'' in \emph{Advances in Neural Information Processing Systems},
  2017, pp. 4077--4087.

\bibitem{li2017crowdsourced}
G.~Li, Y.~Zheng, J.~Fan, J.~Wang, and R.~Cheng, ``Crowdsourced data management:
  Overview and challenges,'' in \emph{ACM International Conference on
  Management of Data}, 2017, pp. 1711--1716.

\bibitem{wang2012crowder}
J.~Wang, T.~Kraska, M.~J. Franklin, and J.~Feng, ``Crowder: crowdsourcing
  entity resolution,'' \emph{VLDB Endowment}, vol.~5, no.~11, pp. 1483--1494,
  2012.

\bibitem{wang2013leveraging}
J.~Wang, G.~Li, T.~Kraska, M.~J. Franklin, and J.~Feng, ``Leveraging transitive
  relations for crowdsourced joins,'' in \emph{ACM SIGMOD International
  Conference on Management of Data}, 2013, pp. 229--240.

\bibitem{mozafari2014scaling}
B.~Mozafari, P.~Sarkar, M.~Franklin, M.~Jordan, and S.~Madden, ``Scaling up
  crowd-sourcing to very large datasets: a case for active learning,''
  \emph{VLDB Endowment}, vol.~8, no.~2, pp. 125--136, 2014.

\bibitem{yoo2019learning}
D.~Yoo and I.~S. Kweon, ``Learning loss for active learning,'' in \emph{IEEE
  Conference on Computer Vision and Pattern Recognition}, 2019, pp. 93--102.

\bibitem{tu2020crowdwt}
J.~Tu, G.~Yu, J.~Wang, C.~Domeniconi, M.~Guo, and X.~Zhang, ``Crowdwt:
  crowdsourcing via joint modeling of workers and tasks,'' \emph{ACM
  Transactions on Knowledge Discovery from Data}, vol.~99, no.~1, pp. 1--24,
  2020.

\bibitem{yu2020active}
G.~Yu, J.~Tu, J.~Wang, C.~Domeniconi, and X.~Zhang, ``Active multilabel crowd
  consensus,'' \emph{IEEE Transactions on Neural Networks and Learning
  Systems}, vol.~99, no.~1, 2020.

\bibitem{marcus2012counting}
A.~Marcus, D.~Karger, S.~Madden, R.~Miller, and S.~Oh, ``Counting with the
  crowd,'' \emph{VLDB Endowment}, vol.~6, no.~2, pp. 109--120, 2012.

\bibitem{zheng2018dlta}
L.~Zheng and L.~Chen, ``Dlta: A framework for dynamic crowdsourcing
  classification tasks,'' \emph{IEEE Transactions on Knowledge and Data
  Engineering}, vol.~31, no.~5, pp. 867--879, 2018.

\bibitem{tong2018dynamic}
Y.~Tong, L.~Wang, Z.~Zhou, L.~Chen, B.~Du, and J.~Ye, ``Dynamic pricing in
  spatial crowdsourcing: A matching-based approach,'' in \emph{ACM SIGMOD
  International Conference on Management of Data}, 2018, pp. 773--788.

\bibitem{tong2018slade}
Y.~Tong, L.~Chen, Z.~Zhou, H.~V. Jagadish, L.~Shou, and W.~Lv, ``Slade: A smart
  large-scale task decomposer in crowdsourcing,'' \emph{IEEE Transactions on
  Knowledge and Data Engineering}, vol.~30, no.~8, pp. 1588--1601, 2018.

\bibitem{yarowsky1995unsupervised}
D.~Yarowsky, ``Unsupervised word sense disambiguation rivaling supervised
  methods,'' in \emph{Annual Meeting of the Association for Computational
  Linguistics}, 1995, pp. 189--196.

\bibitem{xie2020self}
Q.~Xie, M.-T. Luong, E.~Hovy, and Q.~V. Le, ``Self-training with noisy student
  improves imagenet classification,'' in \emph{IEEE Conference on Computer
  Vision and Pattern Recognition}, 2020, pp. 10\,687--10\,698.

\bibitem{fang2014active}
M.~Fang, J.~Yin, and D.~Tao, ``Active learning for crowdsourcing using
  knowledge transfer,'' in \emph{AAAI Conference on Artificial Intelligence},
  2014, pp. 1809--1815.

\bibitem{korycki2017combining}
{\L}.~Korycki and B.~Krawczyk, ``Combining active learning and self-labeling
  for data stream mining,'' in \emph{International Conference on Computer
  Recognition Systems}, 2017, pp. 481--490.

\bibitem{zhang2017improving}
J.~Zhang, V.~S. Sheng, T.~Li, and X.~Wu, ``Improving crowdsourced label quality
  using noise correction,'' \emph{IEEE Transactions on Neural Networks and
  Learning Systems}, vol.~29, no.~5, pp. 1675--1688, 2017.

\bibitem{zhang2018ensemble}
J.~Zhang, M.~Wu, and V.~S. Sheng, ``Ensemble learning from crowds,'' \emph{IEEE
  Transactions on Knowledge and Data Engineering}, vol.~31, no.~8, pp.
  1506--1519, 2018.

\bibitem{zhang2018mtl}
Y.~Zhang and Q.~Yang, ``An overview of multi-task learning,'' \emph{Nature
  Science Review}, vol.~5, no.~1, pp. 30--43, 2018.

\bibitem{zheng2017truth}
Y.~Zheng, G.~Li, Y.~Li, C.~Shan, and R.~Cheng, ``Truth inference in
  crowdsourcing: Is the problem solved?'' \emph{Proceedings of the VLDB
  Endowment}, vol.~10, no.~5, pp. 541--552, 2017.

\bibitem{dawid1979maximum}
A.~P. Dawid and A.~M. Skene, ``Maximum likelihood estimation of observer
  error-rates using the em algorithm,'' \emph{Journal of the Royal Statistical
  Society: Series C (Applied Statistics)}, vol.~28, no.~1, pp. 20--28, 1979.

\bibitem{whitehill2009whose}
J.~Whitehill, T.-f. Wu, J.~Bergsma, J.~Movellan, and P.~Ruvolo, ``Whose vote
  should count more: Optimal integration of labels from labelers of unknown
  expertise,'' in \emph{Advances in Neural Information Processing Systems},
  2009, pp. 2035--2043.

\bibitem{ipeirotis2010quality}
P.~G. Ipeirotis, F.~Provost, and J.~Wang, ``Quality management on amazon
  mechanical turk,'' in \emph{Proceedings of the ACM SIGKDD workshop on human
  computation}, 2010, pp. 64--67.

\bibitem{guo2012so}
S.~Guo, A.~Parameswaran, and H.~Garcia-Molina, ``So who won? dynamic max
  discovery with the crowd,'' in \emph{ACM SIGMOD International Conference on
  Management of Data}, 2012, pp. 385--396.

\bibitem{karger2011iterative}
D.~R. Karger, S.~Oh, and D.~Shah, ``Iterative learning for reliable
  crowdsourcing systems,'' in \emph{Advances in Neural Information Processing
  Systems}, 2011, pp. 1953--1961.

\bibitem{zhang2013reducing}
C.~J. Zhang, L.~Chen, H.~V. Jagadish, and C.~C. Cao, ``Reducing uncertainty of
  schema matching via crowdsourcing,'' \emph{VLDB Endowment}, vol.~6, no.~9,
  pp. 757--768, 2013.

\bibitem{vinyals2016matching}
O.~Vinyals, C.~Blundell, T.~Lillicrap, D.~Wierstra \emph{et~al.}, ``Matching
  networks for one shot learning,'' in \emph{Advances in Neural Information
  Processing Systems}, 2016, pp. 3630--3638.

\bibitem{griffin2007caltech}
G.~Griffin, A.~Holub, and P.~Perona, ``Caltech-256 object category dataset,''
  2007.

\bibitem{wah2011caltech}
C.~Wah, S.~Branson, P.~Welinder, P.~Perona, and S.~Belongie, ``The caltech-ucsd
  birds-200-2011 dataset,'' 2011.

\end{thebibliography}
\end{document}